\documentclass[11pt]{article}

\usepackage[final]{acl}

\usepackage{times}
\usepackage{latexsym}
\usepackage{amsmath}
\usepackage{longtable}
\usepackage{tabularx}
\usepackage{multirow}
\usepackage{makecell}
\usepackage{booktabs}
\usepackage{capt-of}

\usepackage[T1]{fontenc}

\usepackage[utf8]{inputenc}

\usepackage{microtype}

\usepackage{inconsolata}

\usepackage{graphicx}

\title{EGT-KG: Evidence-Grounded Typed KG Retrieval for Practical Scientific QA with Small Language Models}

\author{
{\bf Muran Yu$^{1}$, Jiechao Gao$^{1}$, Yuandong Pan$^{1}$, Barney H. Miao$^{1,2}$}\\
{\bf Andrew C. Lesh$^{1}$, Kincho H. Law$^{1}$, Jie Wang$^{1}$, Michael D. Lepech$^{1}$}\\
$^{1}$Department of Civil and Environmental Engineering, Stanford University\\
$^{2}$Department of Civil Engineering, University of Calgary\\
{\tt\small \{yumuran, jiechao, ydpan, barneym, aclesh, law, jiewang, mlepech\}@stanford.edu}
}

\begin{document}
\maketitle
\begin{abstract}
For emerging scientific research domains, local Small Language Models (SLMs) are becoming more attractive, as they offer stronger privacy control and more stable deployment pipelines than Large Language Models. 
However, in practice, scientific question-answering on SLMs often operates under inevitable constraints: small literature collections, fragmented evidence, limited context window and reasoning abilities. We propose the Evidence-Grounded Typed Knowledge Graph (EGT-KG), a retrieval framework to improve information retrieval with local SLMs. We assessed three question-answering settings: a vanilla Retrieval-Augmented Generation (RAG) workflow and two EGT-KG workflows: an automatically generated relation schema (AS) and an expert-defined relation schema (ES). Our experiments were evaluated with a six-dimensional evaluation framework (S3CRF: Soundness, Correctness, Completeness, Conciseness, Relevance, Fluency) on a Biopolymer-bound Soil Composite literature benchmark, showing that EGT-KG outperforms the vanilla RAG method in most settings, with the best improvement from \textit{llama3:8b}: a Final Score of 70.37 (+14.67\%) and 68.82 (+12.14\%) by AS/ES EGT-KG variants.
\end{abstract}

\section{Introduction}

The rapid evolution of Large Language Models (LLMs) has changed the paradigm in scientific knowledge discovery and human-AI collaboration, including scientific question-answering~\citep{zheng2025automation}. However, the deployment of frontier LLMs for specialized research workflows can raise concerns about data privacy, operational costs, and inference latency \citep{pan2025cost, cai_latency}. Therefore, Small Language Models (SLMs) have become increasingly attractive for domain-specific applications in constrained environments~\citep{wang_smlsurvey,lu2024small}. 
Yet, SLMs are often unreliable for specialized scientific question-answering because they have limited internal knowledge, weaker reasoning ability, and lower tolerance for noisy context than frontier LLMs \citep{allen2024physics, lee_noise, mallen_not}. This challenge is particularly critical in emerging research domains, where knowledge is fragmented across a small set of knowledge sources and is not well represented in pretraining corpora.

Retrieval-Augmented Generation (RAG) \citep{lewis_rag} can ground local models in external documents, but conventional dense retrieval is often insufficient for scientific QA. Scientific questions may depend on relations among materials, procedures, conditions, observations, and results, while dense RAG mainly retrieves chunks by semantic similarity. As a result, retrieved passages may be broadly relevant but still miss the specific evidence needed for faithful answering.

Graph-based RAG methods address part of this issue by organizing information into entities and relations \citep{edge_graphrag}. However, general knowledge graphs may lack timeliness and fine-grained domain knowledge \citep{ding2024automated}, and automatically constructed graphs can introduce noisy or ambiguous relations \citep{schafer2024_bio_auto,zhuang2025linearrag}. Moreover, converting scientific narratives into triples may discard experimental context, conditional qualifiers, and quantitative details that are necessary for reasoning \citep{pujara-etal-2017-sparsity}.

We propose the Evidence-Grounded Typed KG (EGT-KG) framework to improve RAG with local SLMs in scientific question-answering settings. Our method extends graph-based retrieval with two principles: First, graph structures should guide retrieval rather than replace the original evidence. Second, the relation structure should be informative enough to be useful in retrieving. Thus, we construct a reified knowledge graph with evidence nodes and classified relations, allowing the system to use relation labels as a retrieval signal.

We evaluate our framework on Biopolymer-bound Soil Composite (BSC)-centered materials QA benchmark built from a 30-paper corpus. The knowledge of BSC is distributed across a compact but specialized literature collection, making it a realistic test case for small-corpus scientific QA.


We compared a vanilla RAG workflow with two EGT-KG variants: an automatically generated schema (AS) and an expert-designed schema (ES). The results show that EGT-KG enhanced retrieval improves answer quality for most cases, particularly in factual retrieval questions. The core contributions of this paper are summarized as follows:
First, to enhance RAG in limited resource settings, we propose an Evidence-Grounded Typed KG framework that combines relation classification with provenance-aware evidence nodes for scientific question-answering with local SLMs. Second, we performed a controlled comparison on the BSC-centered benchmark between different workflows, systematically studying the trade-off between schema granularity and answer quality. Third, we introduce an S3CRF evaluation setting that combines multidimensional scoring (\textbf{S}oundness, \textbf{C}orrectness, \textbf{C}ompleteness, \textbf{C}onciseness, \textbf{R}elevance, and \textbf{F}luency) and retrieval performance analysis, illustrating how answer quality is affected by the six criteria, query expansion, fallback rate, and evidence-window usage.

\section{Related Works}

\textbf{Retrieval.} Retrieval-Augmented Generation (RAG) is a widely adopted approach to improving question-answering by providing external knowledge that is not stored in the model's parameters \citep{rag}. It can improve factual grounding and reduce hallucinations, demonstrating its potential for question-answering in professional domains such as medicine, education, and law \citep{rag_survey}. However, dense vector representations can lead to irrelevant documents, potentially ignoring explicit structured dependencies, properties, conditions, or procedures. In addition, computational overhead, memory usage, and RAG poisoning are of concern in the implementation of RAG system \citep{gupta2024comprehensive, greshake_not_what, liu2023prompt, liu2024formalizing}. Therefore, a cost-efficient, stable, and secure RAG system is demanded for sensitive tasks that require high precision. 

\textbf{Graph-RAG.} Knowledge graph-enhanced retrieval methods mitigate the limitations of RAG by providing global descriptions and the hierarchical community structure of the corpus \citep{graph_rag}. Graph-RAG has shown that the graph structure can help multi-hop evidence discovery, avoiding the loss of target information in a single semantically similar chunk \citep{procko2024graph}. However, current graph-based approaches compress the original text, which can potentially lead to information loss and further degrade response generation. Additionally, it also tends to utilize a large portion of triples in the knowledge graph. Without constraints, users may face context explosion and high reasoning costs \citep{xiang2025_explode}. \citet{guo_lightrag} proposed LightRAG to integrate a graph-based indexing approach to improve the performance of RAG in both efficiency and comprehension in information retrieval. 
However, the final passage that is passed to the answer model is still pre-processed by the language model, potentially introducing unverified noise. 


\textbf{Relation Classification.} Relation classification is a well-studied research area in NLP since the last decade. Methods such as feature-based models, kernel-based models, neural-network models, and transformer encoders are utilized to assign a semantic relation to two nominals \citep{kambhatla2004relation_2, bunescu2005relation_3, socher2012relation_4, huang2020relation_1}. However, these applications in relation classification focus on processing original text information rather than the relation node from the extracted triple in knowledge graph. \citet{yu2020relationship} proposed a relationship extraction method for construction of domain knowledge graphs using existing reference material from Wikipedia. Along with other semi-supervised relation classification methods \citep{agichtein2000snowball,ravichandran2002learning, pantel2006espresso}, these classification processes require the collection of additional corpora from the internet. Applying these methods to our framework will introduce additional computational costs, which is stressful for local SLMs.

\section{Methodology}

\begin{figure*}[!t]
    \centering
    \includegraphics[width=\textwidth]{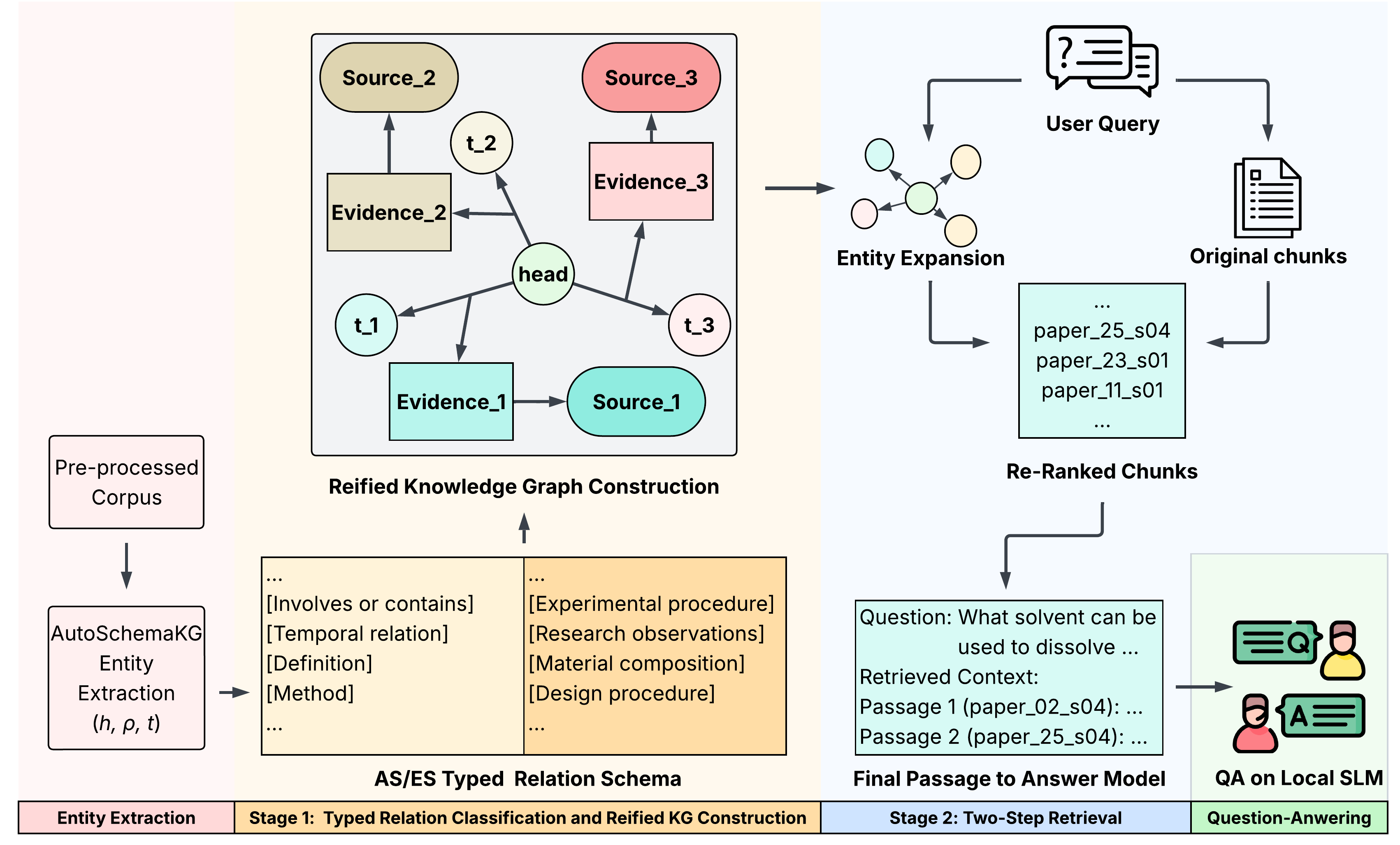}
    \caption{Overview of the EGT-KG framework. The framework consists of four main sections: entity extraction, typed relation classification and reified KG construction, two-step retrieval with re-ranking, and final QA.
    }
    \label{full}
\end{figure*}

We propose an Evidence-Grounded Typed Knowledge Graph retrieval framework for domain-specific question-answering on SLMs. This framework aims to improve the retrieval quality, where the final answer will be grounded on original evidence nodes rather than pre-processed information.

The complete workflow for our framework is organized as follows. In the first stage, document chunks are processed to extract raw triples with short supporting evidence spans. Then, the extracted free-form relation labels are mapped into the two predefined typed schemas accordingly (AS and ES). These typed triples will be used to construct a reified knowledge graph with evidence nodes converted from raw evidence spans, linking the triples to unprocessed evidence text and source chunks. In the second stage, the reified knowledge graph guides query expansion and chunk candidates retrieval. Then the re-ranked chunk candidates and evidence window are passed to the answer model. Fig~\ref{full} provides an overview of the framework structure.


\noindent\textbf{3.1 Graph triple extraction}

\noindent The corpus set is built on a 30-paper corpus centered on Biopolymer-bound Soil Composite (BSC) and related emerging biopolymer-based materials. It includes 9 papers directly related to the evaluation questions, 9 BSC papers not directly related to the questions, 9 materials science papers similar to BSC, and 3 general reference papers. This corpus has three unique properties that make it a suitable test case. First, biopolymer-bound soil composites are an emerging material. So, a language model is unlikely to have much knowledge about them parametrically, which can be confirmed by baseline performance in Table \ref{tab:main_results}. Therefore, the improvement of the framework can be attributed to retrieval quality from the corpus. Second, researchers who work on these materials designed the evaluation questions, so the questions are practically meaningful and cover aspects such as material behavior, material properties, and end-of-life treatment. Third, that same expertise made an expert-defined schema that can better capture useful information in KG. Appendix~\ref{app:bsc} introduces the material and Table~\ref{tab:bsc_papers} lists the 30 papers.

We divided the collected corpus by document structure, preserving metadata such as section headings, resulting in 884 chunks averaging 1,387 characters. Thus, the retrieved evidence span can be traced back to its original document and local position. To ensure that the downstream comparisons are not confounded by different triple extraction passes, we perform triple extraction only once over the fixed chunk corpus and reuse the raw outputs for both EGT-KG variants.

Formally, let $C$ denote the whole chunk set, and each $c \in C$ carries the metadata $M(c)$ including chunk ID, paper ID, source ID, title, chunk length, and original text.

We utilize \textbf{AutoSchemaKG} \citep{chen_autokg} as a triple-extraction engine. Specifically, we use AutoSchemaKG's extraction utilities to process each chunk in $C$ and record structural information. 

As a key feature, the extraction output preserves a head entity, a tail entity, a relation, and a short supporting evidence span with localization information rather than a processed summary or description of the triple. Thus, for each chunk $c\in C$, the extractor returns a set of raw triples $T$ that can be represented as: $T(c) =\{h,\rho| e,t\}$, where $h$ and $t$ denote the head and tail entity, $\rho$ is the free-form relation label, and $e$ is a short evidence span that can be traced back to the source chunk. The evidence span $e$ is appended with the relation $\rho$ to be converted to an evidence node for graph construction, which enables Evidence Window slicing that can offer 22\% context reduction with no cost in performance. The triple set $T$ will be used by both AS/ES variants in subsequent experiments.

\begin{table*}[t]
\centering
\small
\begin{tabularx}{\textwidth}{p{0.24\textwidth} X}
\toprule
\textbf{Relation Type} & \textbf{Composition} \\
\midrule
\makecell[tl]{AS schema\\ (automatically generated)}
& Involves or Contains; Causes or Influences; Temporal Relation; Logical or Preposition; Research or Observation; Method; Definition; Production; Spatial Relation. \\
\midrule
\makecell[tl]{ES schema\\ (expert designed)}
& Causes or Influences; Definition; Manufacture Procedure; Temporal Relation; Experimental Methods; Logical Reasoning; Research Observations; Experimental Procedures; Material Composition; Design Procedure; Biopolymer Properties. \\
\bottomrule
\end{tabularx}
\caption{Typed relation schemas for knowledge graph construction.}
\label{typed_relation}
\end{table*}

\noindent\textbf{3.2 Typed relation classification}

\noindent After triple extraction, we transform relation labels into a set of semantically meaningful relation types. We classify each distinct label once into a schema type or discard it, then apply the resulting label-to-type map to every triple with that label, so identical labels are always typed identically. During this process, we restrict the size of the knowledge graph by trimming noisy, inconsistent, and irrelevant triples for retrieval. 

The two type schemas are designed to study how relation granularity itself can alter the process of query expansion and quality of responses. As shown in Table~\ref{typed_relation}, the AS schema is automatically generated by answer models based on specific instructions for the BSC study, and the generation prompt is given in Appendix~\ref{app:prompts}. We selected the most prominent types and applied them to all AS EGT-KG variant studies. The ES schema was defined by two materials scientists with several years of research experience in relevant materials. They reviewed and identified the relation labels produced, and grouped them into 11 types. The AS–ES contrast therefore isolates the value of expert input, and the ES schema can potentially be expanded to broader materials science domains.

Formally, let $\phi$ denote the relation typing function, and $\tau$ denote the type schema: $\phi(T(c)) = (h,\tau_n|\rho| e,t)$ ,where $\tau_1$ is the AS schema and $\tau_2$ is the ES schema. Applying the relation typing function $\phi$ will classify each of the extracted triples $t$ into a typed relation instance and be appended to the relation label. Thus, by differentiating relation types, our framework provides a high-level index to guide graph traversal and constrains query expansion, forcing the retrieval process to operate only on the most stable semantic categories. 

\noindent\textbf{3.3 Reification and provenance-aware knowledge graph construction} 

\noindent \citet{reinfication} states that although current methods of knowledge graph embedding perform well on total and functional relations, features such as partiality or non-functionality of relations are difficult to model correctly. So, they introduced the concept of Reification, which represents relations directly as objects in the embedding space rather than as geometric transformations from subject to object. For our framework, the purpose of reification is to preserve each relation instance as a same-level object linked to both its semantic arguments and its original support text. Therefore, rather than representing knowledge as direct entity-to-entity edges only, the reified knowledge graph stores two kinds of additional nodes besides the head and tail. The first is the evidence nodes, which contain the supporting evidence span $e$, and the second is the source nodes, which store the location of the evidence span in the original document. This reified design is central to our framework because it allows graph retrieval to remain structurally expressive while preserving the textual provenance required for faithful answer generation. This structure is especially important in scientific question-answering, where relation abstraction or description alone may omit crucial conditions, quantitative ranges, or contextual qualifiers in the original text information. So far, the first stage is completed, and the overview of stage 1 is shown in Appendix~\ref{app:pipeline_details}. 

\noindent\textbf{3.4 Two-step retrieval with query expansion and re-ranking
} 

\noindent At the beginning of the second stage, our EGT-KG framework follows a two-step retrieval strategy. In the first step, to improve recall, the answer model retrieves chunk candidates by using the reified knowledge graph to guide query expansion. In the second step, to improve precision, the retrieved chunk candidates are re-ranked according to their evidence-level relevance. 
The two-step retrieval design is guided by the hypothesis of our work, in which query expansion helps find the most relevant neighbor entities, while chunk candidates re-ranking helps retrieve the correct final passages.

In practice, the two-step retrieval works as follows. In the first step, given a user query $q$, we identify seed entities in KG through query-entity matching. Next, we perform a one-hop query expansion around the seed entities to collect related entities. The first-step retrieval score ($S_1$) is the weighted sum of the original-query score $S_{ori}$ and the expanded-query relevance score $S_{exp}$, allowing the retrieval to benefit from both global and local views. This score will be used to select top-k chunk candidates. Formally, for a chunk candidate $c$ and weight term $\alpha$:
\begin{equation}
    S_1(c|q) = \alpha S_{ori}(c|q) + (1-\alpha)S_{exp}(c|q)
    \label{eq:s1}
\end{equation}

For the second step, the top-k chunk candidates collected in step 1 are re-ranked by comparing the semantic similarity between the query and the content of the evidence nodes. The re-ranking process provides more precise support evidence within the retrieved material and determines if the final answer context should be constructed from localized evidence window or from full chunk fallback. Specifically, let $S_{evi}$ denote the evidence-level score, and $\lambda$ denote the weight for step 2, the re-ranking score $S_{2}$ is defined as:
\begin{equation}
    S_{2}(c,e|q) = \lambda S_1(c|q) + (1-\lambda)S_{evi}(e|q)
    \label{eq:s2}
\end{equation}

For a complete illustration, Figure~\ref{stage2} in Appendix~\ref{app:pipeline_details} provides a detailed example. 

\noindent\textbf{3.5 Evidence-grounded answer generation}

\noindent After two-step retrieval and re-ranking, the system constructs the final context passage from either top-k evidence windows or full-chunk fallbacks. 
All compared workflows, including the vanilla RAG and two EGT-KG variants, share the same answer-generation policy, instructing the answer model to answer the question directly with no unsupported inference, and explicitly acknowledging reference information insufficiency when retrieved chunks do not contain enough information. 
Fig~\ref{stage2} in Appendix~\ref{app:pipeline_details} shows the overview of the second stage.

Formally, given a question $q$ and a final context passage $P_q$, the shared evidence-based prompt $\pi$, the answer model produces:
\begin{equation}
    answer = SLM(q,P_q,\pi)
\end{equation}

\noindent\textbf{3.6 Evaluation with S3CRF}

\noindent All generated responses are evaluated under a unified LLM-as-judge workflow. 
Due to the unique nature of our problem, current benchmarks are not best suited to evaluate emerging research domains with limited reference sources and open-ended questions. We adopt the S3CRF framework, scoring answers along six dimensions: soundness, correctness, completeness, conciseness, relevance, and fluency. The detailed evaluation criteria are shown in Table~\ref{rubric-vertical}. 
Domain experts in BSC were involved in validating the benchmark design and evaluation criteria to improve the credibility of the assessment.

The six-dimensional scores ($s_n$) are combined into a weighted Final Score ($S_{final}$) to reflect the benchmark's emphasis on scientific question-answering demands
, the overall Final Score with weights $w_n$ is computed as: $S_{final}= \sum (w_n s_n)$.

\section{Results}

\begin{table*}[!t]
\centering
\small
\setlength{\tabcolsep}{4pt}
\begin{tabular}{lcccccccccc}
\toprule
& \multicolumn{2}{c}{Gemma2:9b}
& \multicolumn{2}{c}{Llama3:8b}
& \multicolumn{2}{c}{Llama3.1:8b}
& \multicolumn{2}{c}{Mistral:7b}
& \multicolumn{2}{c}{Qwen2.5:7b} \\
\cmidrule(lr){2-3} \cmidrule(lr){4-5} \cmidrule(lr){6-7} \cmidrule(lr){8-9} \cmidrule(lr){10-11}
& gpt-4o & gemini
& gpt-4o & gemini
& gpt-4o & gemini
& gpt-4o & gemini
& gpt-4o & gemini \\
\midrule
Baseline    & 35.96 & 24.60 & 28.02 & 24.80 & 29.19 & 24.95 & 28.65 & 22.29 & 33.51 & 24.43 \\
Vanilla RAG   & 66.76 & 70.58 & 61.37 & 62.44 & 84.30 & 77.33 & 59.32 & 77.55 & \textbf{86.01} & 78.09 \\
EGT-KG(AS)  & 72.22 & 72.62 & \textbf{70.37} & 66.11 & 83.75 & 78.44 & 60.34 & \textbf{77.78} & 82.82 & \textbf{80.55} \\
EGT-KG(ES) & \textbf{73.12} & \textbf{74.04} & 68.82 & \textbf{68.21} & \textbf{86.02} & \textbf{81.72} & \textbf{60.57} & 74.23 & 85.20 & 79.91 \\
\bottomrule
\end{tabular}
\caption{Performance of workflows across answer models and judge models.}
\label{tab:main_results}
\end{table*}

\begin{table*}[!t]
\centering
\small
\setlength{\tabcolsep}{4pt}
\begin{tabular}{llccccc}
\toprule
Model & Relation Type & Fallback & Window & AvgExpand & EE & Jaccard\\
\midrule
Gemma2:9b   & AS  & 0.6292 & 0.3708 & 10.40 & 3.57\% & 0.6095\\
Gemma2:9b   & ES & 0.6167 & 0.3833 & 10.00 & 3.83\% & 0.5678\\
\midrule
Llama3:8b   & AS  & 0.7500 & 0.2500 & 14.95 & 1.67\% &0.5961\\
Llama3:8b   & ES & 0.7667 & 0.2333 & 16.90 & 1.38\% &0.5855\\
\midrule
Llama3.1:8b & AS  & 0.6042 & 0.3958 & 17.10 & 2.31\% &0.5762\\
Llama3.1:8b & ES & 0.6375 & 0.3625 & 16.55 & 2.19\% &0.5427\\
\midrule
Mistral:7b  & AS  & 0.7792 & 0.2208 & 13.13 & 1.68\% &0.6186\\
Mistral:7b  & ES & 0.7917 & 0.2083 & 12.80 & 1.63\% &0.6037\\
\midrule
Qwen2.5:7b  & AS  & 0.7167 & 0.2833 & 24.15 & 1.17\% &0.4930\\
Qwen2.5:7b  & ES & 0.7208 & 0.2792 & 25.05 & 1.11\% &0.4847\\
\bottomrule
\end{tabular}

\vspace{0.5\baselineskip}
\caption{Retrieval performance analysis. Expansion efficiency ($EE$) is defined as $\mathrm{WindowRate}/\mathrm{AvgExpanded}$, and the Jaccard index measures the difference between original retrieved and re-ranked chunk candidates.}
\label{tab:retrieval_stats}
\end{table*}

\noindent\textbf{4.1 Experimental comparison setup}

\noindent All compared workflows share the same strict evidence-grounded instruction prompt and the S3CRF multidimensional LLM-as-Judge evaluation framework
. For all experiments, the answer models receive the top-k ($k=6$) retrieved passages to construct the final prompt for each query (Appendix ~\ref{app:params}). We report results for five local answer models evaluated by \textit{gpt-4o} \citep{chatgpt4o} and \textit{gemini-2.5-pro} \citep{gemini25pro} as judge models.

\noindent\textbf{4.2 Overall performance across models}

\noindent Table~\ref{tab:main_results} summarizes the overall performance of baseline, vanilla RAG, and two EGT-KG variants. 
Generally, EGT-KG enhanced retrieval improves answer quality in most cases, but the improvement is clearly model-dependent. The highest absolute values of Final Scores are achieved by \textit{llama3.1:8b} in the benchmark, and \textit{gemma2:9b} demonstrates the most stable improvements from EGT-KG enhanced retrieval. Besides, \textit{llama3:8b} shows the greatest relative improvement compared to the vanilla RAG workflow, while \textit{mistral:7b} and \textit{qwen2.5:7b} exhibit smaller or less stable improvements compared to the other answer models.

The largest relative improvement appears for \textit{llama3:8b} under \textit{gpt-4o} evaluation, where the Final Score increases from 61.37 with vanilla RAG to 70.37 (+14.67\%) with EGT-KG(AS) and 68.82 (+12.14\%) with EGT-KG(ES). For \textit{gemma2:9b}, it outperforms vanilla RAG under both judge models, presenting the most stable improvements as the granularity of the type schema increases. And \textit{llama3.1:8b} reaches the highest overall scores with the ES schema (86.02 by \textit{gpt-4o} and 81.72 by \textit{gemini-2.5-pro}). By contrast, \textit{mistral:7b} and \textit{qwen2.5:7b} show smaller or less stable improvements, indicating that EGT-KG is not uniformly beneficial for all local models.

These results support two conclusions. First, our EGT-KG framework improved scientific question-answering by two-step retrieval from the reified knowledge graph in most settings. Second, generally, a finer-grained type schema can better guide answer models to retrieve the most relevant information from the original text to help with question-answering. However, given the differences in model structure and reasoning abilities, the benefit of additional granularity is conditional rather than universal.

As a supplementary generalization check, we evaluate EGT-KG on QASPER~\citep{qasper} with a cross-paper setting, where the system retrieves from all 281 papers without access to the gold paper ID. As shown in Appendix~\ref{app:qasper}, EGT-KG improves mean Answer F1 over vanilla RAG by 5.9\% on average, with gains on four out of the five base models. We additionally evaluate on HotpotQA. The outcome depends entirely on how large a candidate pool retrieval must search: the framework adds almost nothing when the official distractor setting supplies ten candidates per question, but improves Answer F1 by $+0.142$ (\textit{gemma2:9b}) once the same paragraphs are pooled into a single corpus (Appendix~\ref{app:hotpot}).

\noindent\textbf{4.3 Retrieval performance and error analysis}
\label{performance_analysis}

\noindent To further investigate why the EGT-KG framework can help more in some settings than others, we conducted a retrieval performance analysis in Table~\ref{tab:retrieval_stats}. The fallback rate is the fraction of the final $k$ chunks for which no evidence window could be constructed, and AvgExpand is the mean number of expansion entities produced per question by one-hop traversal.
The most positive cases (\textit{gemma2:9b} and \textit{llama3.1:8b}) show the lowest fallback rates and highest evidence-window utilization ($EE$). By contrast, \textit{mistral:7b} has the highest fallback rates (0.78-0.79) with limited improvements in Final Score. Compared to \textit{gemma2:9b}, \textit{mistral:7b} has a similar Jaccard index but a much higher fallback rate, suggesting that the model successfully re-ranked the original chunk candidates, but failed to extract a useful evidence window. 

Also, by analyzing average expanded ($AvgExpand$) query entities, we found that retrieving more neighbor entities does not always guarantee better results. Compared to other answer models, \textit{qwen2.5:7b} collected the largest number of neighbor entities through one-hop query expansion, expanding to 25.05 entities per question under ES with the lowest expansion efficiency. The lowest Jaccard index demonstrates \textit{qwen2.5:7b} replaced more original chunk candidates than other answer models, indicating that the model failed to extract useful evidence from these new chunk entities, while these underutilized chunk candidates distracted model's attention from the most relevant information.

\section{Discussion}

\noindent\textbf{5.1 EGT-KG enhanced retrieval helps by improving evidence access}

\noindent In general, the experimental results suggest that the main value of the EGT-KG framework lies in improving the precision of textual evidence passed to the answer model. With the same prompt structure, the improvements observed in EGT-KG variants over the vanilla RAG workflow can plausibly be explained by better retrieval quality and evidence localization than by refined prompt engineering. 

To be specific, the reified knowledge graph helps the answer model to identify relevant entities, expand query around structured neighbor entities, and re-rank chunk candidates based on evidence-level matching. Thus, the EGT-KG framework should be treated as a lightweight retrieval booster for SLMs in a limited resources setting. 

\noindent\textbf{5.2 Finer-grained schema is not always better}

\noindent The comparison between the two variants demonstrated that increased granularity does not always lead to better performance. Although some models can consistently benefit from the finer-grained type schema, the coarse AS EGT-KG variants remain more effective for models that cannot fully interpret additional knowledge injected. This result reveals the trade-off between semantic precision and retrieval robustness. Finer schema granularity can surely provide more informative instruction that aligns the original query with reified knowledge graph, but on the other hand, it may introduce sparsity and classification noise, especially for local SLMs, which are typically less robust to noise or weakly aligned context than frontier LLMs. The coarse schema, with more general relation classes defined, may not be as specific as the finer one, but it provides a more stable retrieval path to SLMs.

\noindent\textbf{5.3 Retrieval quality is mainly constrained by evidence localization}

\noindent The retrieval analysis in section~\ref{performance_analysis} indicates that the bottleneck of retrieval quality is evidence localization quality, rather than the number of retrieved chunk candidates. The framework successfully expands the range of retrieved chunk candidates via one-hop query expansion with 30\% of the original chunk candidates replaced after re-ranking. However, the response quality highly depends on whether the answer model can locate a clear evidence window in retrieved chunks. The models with better performance in our framework gained lower fallback rates and higher evidence-window utilization rates, while the weaker models always fall back to full chunk information passing. Thus, for our specific scenario with limited resources, more graph expansion does not automatically improve results. Instead, the model's ability to convert chunk information into grounded evidence should be prioritized when determining whether the expanded retrieval is helpful or noisy.

\section{Conclusion}

We presented EGT-KG, an evidence-grounded typed KG retrieval framework for practical scientific question-answering with local small language models. 
Our findings suggest that the EGT-KG framework improves answer quality over vanilla RAG in most settings, especially on dimensions related to factual correctness, reasoning soundness, completeness, and relevance. 
This observation is especially important for SLMs when scientific QA systems must operate on specialized corpora with fragmented evidence and limited model capacity. Future work will focus on reducing fallback rate, balancing multi-hop expansion with limited model context length, improving evidence window utilization, and testing the framework on more domains of scientific research.


\section{Limitations}

The current study has several limitations. Firstly, the relatively high fallback rate indicates that there is still a large portion of retrieved chunk candidates that cannot be converted into a clear evidence window. Therefore, the potential of our provenance-aware retrieval is not fully exploited. The current results may be improved with a better solution for evidence localization for SLMs. Secondly, our query expansion relies on one-hop neighbor entity expansion. While this choice takes noise management and SLM's limited reasoning ability into consideration, it reduces our framework's capability in handling complex queries that require multi-hop reasoning or cross-document inference. Third, the current validation is conducted on a BSC-centered benchmark. Although our corpus setting is appropriate as a test case for open-ended question-answering with fragmented and specialized scientific knowledge, the present results should not be generalized to a broader scope of research domains without further validation. The additional evaluation on QASPER has partially shown generalization, but its question set frequently uses pronouns instead of precise terminology. In cross-paper settings, the performance of EGT-KG is underestimated because of the nature of the question set. Fourth, our work focuses on improving retrieval for local SLMs rather than on comparing against frontier LLMs as answer generators. Thus, the results should be interpreted as evidence that better retrieval quality can strengthen feasible local SLMs in constrained resource settings, rather than a claim of general superiority over LLMs.
Finally, for the designed open-ended questions, the LLM-as-judge method remains sensitive to judge model preferences. The use of two judges improves robustness, but the current Final Score difference should be treated as a relative performance indicator rather than as a measure of actual scientific usefulness.




\bibliography{custom}

@misc{qasper,
      title={A Dataset of Information-Seeking Questions and Answers Anchored in Research Papers}, 
      author={Pradeep Dasigi and Kyle Lo and Iz Beltagy and Arman Cohan and Noah A. Smith and Matt Gardner},
      year={2021},
      eprint={2105.03011},
      archivePrefix={arXiv},
      primaryClass={cs.CL},
      url={https://arxiv.org/abs/2105.03011}, 
}

@article{miao2024life,
  title={Life cycle assessment and design of LignoBlock: A lignin bound block on the path towards a green transition of the construction industry},
  author={Miao, Barney H and Headrick, Robert J and Li, Zhiye and Spanu, Leonardo and Loftus, David J and Lepech, Michael D},
  journal={Journal of Cleaner Production},
  volume={474},
  pages={143610},
  year={2024},
  publisher={Elsevier}
}

@inproceedings{rag,
author = {Lewis, Patrick and Perez, Ethan and Piktus, Aleksandra and Petroni, Fabio and Karpukhin, Vladimir and Goyal, Naman and K\"{u}ttler, Heinrich and Lewis, Mike and Yih, Wen-tau and Rockt\"{a}schel, Tim and Riedel, Sebastian and Kiela, Douwe},
title = {Retrieval-augmented generation for knowledge-intensive NLP tasks},
year = {2020},
isbn = {9781713829546},
publisher = {Curran Associates Inc.},
address = {Red Hook, NY, USA},
booktitle = {Proceedings of the 34th International Conference on Neural Information Processing Systems},
articleno = {793},
numpages = {16},
location = {Vancouver, BC, Canada},
series = {NIPS '20}
}

@inproceedings{graph_rag,
author = {Fu, Chenhan and Wang, Guoming and Lu, Rongxing and Tang, Siliang},
title = {Global Discovery: A Global Graph-RAG Approach for Query-Focused Multimodal Summarization Across Multiple PDF Papers},
year = {2025},
isbn = {978-981-95-3060-1},
publisher = {Springer-Verlag},
address = {Berlin, Heidelberg},
url = {https://doi.org/10.1007/978-981-95-3061-8_1},
doi = {10.1007/978-981-95-3061-8_1},
booktitle = {Knowledge Science, Engineering and Management: 18th International Conference, KSEM 2025, Macao, China, August 4–7, 2025, Proceedings, Part V},
pages = {1–8},
numpages = {8},
location = {Macao, China}
}

@article{guo_lightrag,
  title={Lightrag: Simple and fast retrieval-augmented generation},
  author={Guo, Zirui and Xia, Lianghao and Yu, Yanhua and Ao, Tu and Huang, Chao},
  journal={arXiv preprint arXiv:2410.05779},
  year={2024}
}

@inproceedings{pujara-etal-2017-sparsity,
    title = "Sparsity and Noise: Where Knowledge Graph Embeddings Fall Short",
    author = "Pujara, Jay  and
      Augustine, Eriq  and
      Getoor, Lise",
    editor = "Palmer, Martha  and
      Hwa, Rebecca  and
      Riedel, Sebastian",
    booktitle = "Proceedings of the 2017 Conference on Empirical Methods in Natural Language Processing",
    month = sep,
    year = "2017",
    address = "Copenhagen, Denmark",
    publisher = "Association for Computational Linguistics",
    url = "https://aclanthology.org/D17-1184/",
    doi = "10.18653/v1/D17-1184",
    pages = "1751--1756"
}

@inproceedings{zheng2025automation,
  title={From automation to autonomy: A survey on large language models in scientific discovery},
  author={Zheng, Tianshi and Deng, Zheye and Tsang, Hong Ting and Wang, Weiqi and Bai, Jiaxin and Wang, Zihao and Song, Yangqiu},
  booktitle={Proceedings of the 2025 Conference on Empirical Methods in Natural Language Processing},
  pages={17744--17761},
  year={2025}
}

@article{pan2025cost,
  title={A Cost-Benefit Analysis of On-Premise Large Language Model Deployment: Breaking Even with Commercial LLM Services},
  author={Pan, Guanzhong and Chodnekar, Vishal and Roy, Abinas and Wang, Haibo},
  journal={arXiv preprint arXiv:2509.18101},
  year={2025}
}

@inproceedings{cai_latency,
author = {Cai, Tianle and Li, Yuhong and Geng, Zhengyang and Peng, Hongwu and Lee, Jason D. and Chen, Deming and Dao, Tri},
title = {MEDUSA: Simple LLM inference acceleration framework with multiple decoding heads},
year = {2024},
publisher = {JMLR.org},
booktitle = {Proceedings of the 41st International Conference on Machine Learning},
articleno = {203},
numpages = {27},
location = {Vienna, Austria},
series = {ICML'24}
}

@article{wang_smlsurvey,
author = {Wang, Fali and Zhang, Zhiwei and Zhang, Xianren and Wu, Zongyu and Mo, TzuHao and Lu, Qiuhao and Wang, Wanjing and Li, Rui and Xu, Junjie and Tang, Xianfeng and He, Qi and Ma, Yao and Huang, Ming and Wang, Suhang},
title = {A Comprehensive Survey of Small Language Models in the Era of Large Language Models: Techniques, Enhancements, Applications, Collaboration with LLMs, and Trustworthiness},
year = {2025},
issue_date = {December 2025},
publisher = {Association for Computing Machinery},
address = {New York, NY, USA},
volume = {16},
number = {6},
issn = {2157-6904},
url = {https://doi.org/10.1145/3768165},
doi = {10.1145/3768165},
journal = {ACM Trans. Intell. Syst. Technol.},
month = nov,
articleno = {145},
numpages = {87}
}

@misc{lewis_rag,
      title={Retrieval-Augmented Generation for Knowledge-Intensive NLP Tasks}, 
      author={Patrick Lewis and Ethan Perez and Aleksandra Piktus and Fabio Petroni and Vladimir Karpukhin and Naman Goyal and Heinrich Küttler and Mike Lewis and Wen-tau Yih and Tim Rocktäschel and Sebastian Riedel and Douwe Kiela},
      year={2021},
      eprint={2005.11401},
      archivePrefix={arXiv},
      primaryClass={cs.CL},
      url={https://arxiv.org/abs/2005.11401}, 
}

@article{edge_graphrag,
  title={From local to global: A graph rag approach to query-focused summarization},
  author={Edge, Darren and Trinh, Ha and Cheng, Newman and Bradley, Joshua and Chao, Alex and Mody, Apurva and Truitt, Steven and Metropolitansky, Dasha and Ness, Robert Osazuwa and Larson, Jonathan},
  journal={arXiv preprint arXiv:2404.16130},
  year={2024}
}

@article{lu2024small,
  title={Small language models: Survey, measurements, and insights},
  author={Lu, Zhenyan and Li, Xiang and Cai, Dongqi and Yi, Rongjie and Liu, Fangming and Zhang, Xiwen and Lane, Nicholas D and Xu, Mengwei},
  journal={arXiv preprint arXiv:2409.15790},
  year={2024}
}

@misc{chen_autokg,
      title={AutoKG: Efficient Automated Knowledge Graph Generation for Language Models}, 
      author={Bohan Chen and Andrea L. Bertozzi},
      year={2023},
      eprint={2311.14740},
      archivePrefix={arXiv},
      primaryClass={cs.CL},
      url={https://arxiv.org/abs/2311.14740}, 
}

@inproceedings{reinfication,
  title={Knowledge graph embeddings with ontologies: reification for representing arbitrary relations},
  author={Leemhuis, Mena and {\"O}z{\c{c}}ep, {\"O}zg{\"u}r L and Wolter, Diedrich},
  booktitle={German Conference on Artificial Intelligence (K{\"u}nstliche Intelligenz)},
  pages={146--159},
  year={2022},
  organization={Springer}
}

@software{chatgpt4o,
  author = {{OpenAI}},
  title = {ChatGPT},
  version = {GPT-4o},
  year = {2024},
  url = {https://chatgpt.com}
}

@software{gemini25pro,
  author = {{Google}},
  title = {Gemini 2.5 Pro},
  year = {2025},
  url = {https://gemini.google.com}
}

@misc{lee_noise,
      title={Lost in the Noise: How Reasoning Models Fail with Contextual Distractors}, 
      author={Seongyun Lee and Yongrae Jo and Minju Seo and Moontae Lee and Minjoon Seo},
      year={2026},
      eprint={2601.07226},
      archivePrefix={arXiv},
      primaryClass={cs.AI},
      url={https://arxiv.org/abs/2601.07226}, 
}

@article{allen2024physics,
  title={Physics of language models: Part 3.3, knowledge capacity scaling laws},
  author={Allen-Zhu, Zeyuan and Li, Yuanzhi},
  journal={arXiv preprint arXiv:2404.05405},
  year={2024}
}

@inproceedings{mallen_not,
  title={When not to trust language models: Investigating effectiveness of parametric and non-parametric memories},
  author={Mallen, Alex and Asai, Akari and Zhong, Victor and Das, Rajarshi and Khashabi, Daniel and Hajishirzi, Hannaneh},
  booktitle={Proceedings of the 61st annual meeting of the association for computational linguistics (volume 1: Long papers)},
  pages={9802--9822},
  year={2023}
}

@article{ding2024automated,
  title={Automated construction of theme-specific knowledge graphs},
  author={Ding, Linyi and Zhou, Sizhe and Xiao, Jinfeng and Han, Jiawei},
  journal={arXiv preprint arXiv:2404.19146},
  year={2024}
}

@article{
schafer2024_bio_auto,
author = {Henning Schäfer  and Ahmad Idrissi-Yaghir  and Kamyar Arzideh  and Hendrik Damm  and Tabea M.G. Pakull  and Cynthia S. Schmidt  and Mikel Bahn  and Georg Lodde  and Elisabeth Livingstone  and Dirk Schadendorf  and Felix Nensa  and Peter A. Horn  and Christoph M. Friedrich },
title = {BioKGrapher: Initial evaluation of automated knowledge graph construction from biomedical literature},
journal = {Computational and Structural Biotechnology Journal},
volume = {24},
number = {},
pages = {639-660},
year = {2024},
doi = {10.1016/j.csbj.2024.10.017},
URL = {https://spj.science.org/doi/abs/10.1016/j.csbj.2024.10.017},
eprint = {https://spj.science.org/doi/pdf/10.1016/j.csbj.2024.10.017}}

@article{zhuang2025linearrag,
  title={Linearrag: Linear graph retrieval augmented generation on large-scale corpora},
  author={Zhuang, Luyao and Chen, Shengyuan and Xiao, Yilin and Zhou, Huachi and Zhang, Yujing and Chen, Hao and Zhang, Qinggang and Huang, Xiao},
  journal={arXiv preprint arXiv:2510.10114},
  year={2025}
}

@inproceedings{rag_survey,
author = {Fan, Wenqi and Ding, Yujuan and Ning, Liangbo and Wang, Shijie and Li, Hengyun and Yin, Dawei and Chua, Tat-Seng and Li, Qing},
title = {A Survey on RAG Meeting LLMs: Towards Retrieval-Augmented Large Language Models},
year = {2024},
isbn = {9798400704901},
publisher = {Association for Computing Machinery},
address = {New York, NY, USA},
url = {https://doi.org/10.1145/3637528.3671470},
doi = {10.1145/3637528.3671470},
booktitle = {Proceedings of the 30th ACM SIGKDD Conference on Knowledge Discovery and Data Mining},
pages = {6491–6501},
numpages = {11},
location = {Barcelona, Spain},
series = {KDD '24}
}

@article{gupta2024comprehensive,
  title={A comprehensive survey of retrieval-augmented generation (rag): Evolution, current landscape and future directions},
  author={Gupta, Shailja and Ranjan, Rajesh and Singh, Surya Narayan},
  journal={arXiv preprint arXiv:2410.12837},
  year={2024}
}

@inproceedings{liu2024formalizing,
  title={Formalizing and benchmarking prompt injection attacks and defenses},
  author={Liu, Yupei and Jia, Yuqi and Geng, Runpeng and Jia, Jinyuan and Gong, Neil Zhenqiang},
  booktitle={33rd USENIX Security Symposium (USENIX Security 24)},
  pages={1831--1847},
  year={2024}
}

@misc{liu2023prompt,
      title={Prompt Injection attack against LLM-integrated Applications}, 
      author={Yi Liu and Gelei Deng and Yuekang Li and Kailong Wang and Zihao Wang and Xiaofeng Wang and Tianwei Zhang and Yepang Liu and Haoyu Wang and Yan Zheng and Leo Yu Zhang and Yang Liu},
      year={2025},
      eprint={2306.05499},
      archivePrefix={arXiv},
      primaryClass={cs.CR},
      url={https://arxiv.org/abs/2306.05499}, 
}

@inproceedings{greshake_not_what,
author = {Greshake, Kai and Abdelnabi, Sahar and Mishra, Shailesh and Endres, Christoph and Holz, Thorsten and Fritz, Mario},
title = {Not What You've Signed Up For: Compromising Real-World LLM-Integrated Applications with Indirect Prompt Injection},
year = {2023},
isbn = {9798400702600},
publisher = {Association for Computing Machinery},
address = {New York, NY, USA},
url = {https://doi.org/10.1145/3605764.3623985},
doi = {10.1145/3605764.3623985},
booktitle = {Proceedings of the 16th ACM Workshop on Artificial Intelligence and Security},
pages = {79–90},
numpages = {12},
location = {Copenhagen, Denmark},
series = {AISec '23}
}

@article{xiang2025_explode,
  title={When to use graphs in rag: A comprehensive analysis for graph retrieval-augmented generation},
  author={Xiang, Zhishang and Wu, Chuanjie and Zhang, Qinggang and Chen, Shengyuan and Hong, Zijin and Huang, Xiao and Su, Jinsong},
  journal={arXiv preprint arXiv:2506.05690},
  year={2025}
}

@inproceedings{procko2024graph,
  title={Graph retrieval-augmented generation for large language models: A survey},
  author={Procko, Tyler Thomas and Ochoa, Omar},
  booktitle={2024 Conference on AI, science, engineering, and technology (AIxSET)},
  pages={166--169},
  year={2024},
  organization={IEEE}
}

@article{huang2020relation_1,
  title={Relation classification via knowledge graph enhanced transformer encoder},
  author={Huang, Wenti and Mao, Yiyu and Yang, Zhan and Zhu, Lei and Long, Jun},
  journal={Knowledge-based systems},
  volume={206},
  pages={106321},
  year={2020},
  publisher={Elsevier}
}

@inproceedings{kambhatla2004relation_2,
  title={Combining lexical, syntactic, and semantic features with maximum entropy models for information extraction},
  author={Kambhatla, Nanda},
  booktitle={Proceedings of the ACL interactive poster and demonstration sessions},
  pages={178--181},
  year={2004}
}

@inproceedings{bunescu2005relation_3,
  title={A shortest path dependency kernel for relation extraction},
  author={Bunescu, Razvan and Mooney, Raymond},
  booktitle={Proceedings of human language technology conference and conference on empirical methods in natural language processing},
  pages={724--731},
  year={2005}
}

@inproceedings{socher2012relation_4,
  title={Semantic compositionality through recursive matrix-vector spaces},
  author={Socher, Richard and Huval, Brody and Manning, Christopher D and Ng, Andrew Y},
  booktitle={Proceedings of the 2012 joint conference on empirical methods in natural language processing and computational natural language learning},
  pages={1201--1211},
  year={2012}
}

@article{yu2020relationship,
  title={A relationship extraction method for domain knowledge graph construction},
  author={Yu, Haoze and Li, Haisheng and Mao, Dianhui and Cai, Qiang},
  journal={World Wide Web},
  volume={23},
  number={2},
  pages={735--753},
  year={2020},
  publisher={Springer}
}

@inproceedings{agichtein2000snowball,
  title={Snowball: Extracting relations from large plain-text collections},
  author={Agichtein, Eugene and Gravano, Luis},
  booktitle={Proceedings of the fifth ACM conference on Digital libraries},
  pages={85--94},
  year={2000}
}

@inproceedings{pantel2006espresso,
  title={Espresso: Leveraging generic patterns for automatically harvesting semantic relations},
  author={Pantel, Patrick and Pennacchiotti, Marco},
  booktitle={Proceedings of the 21st international conference on computational linguistics and 44th annual meeting of the Association for Computational Linguistics},
  pages={113--120},
  year={2006}
}

@inproceedings{ravichandran2002learning,
  title={Learning surface text patterns for a question answering system},
  author={Ravichandran, Deepak and Hovy, Eduard},
  booktitle={Proceedings of the 40th Annual meeting of the association for Computational Linguistics},
  pages={41--47},
  year={2002}
}

@article{bsc_1,
  title={On designing biopolymer-bound soil composites (BSC) for peak compressive strength},
  author={Rosa, Isamar and Roedel, Henning and Allende, Maria I and Lepech, Michael D and Loftus, David J},
  journal={Journal of Renewable Materials},
  volume={8},
  number={8},
  pages={845--861},
  year={2020},
  publisher={Tech Science Press}
}

@article{bsc_2,
  title={Development of biopolymer composites using lignin: A sustainable technology for fostering a green transition in the construction sector},
  author={Miao, Barney H and Headrick, Robert J and Li, Zhiye and Spanu, Leonardo and Loftus, David J and Lepech, Michael D},
  journal={Cleaner Materials},
  volume={14},
  pages={100279},
  year={2024},
  publisher={Elsevier}
}

@article{bsc_3,
  title={Sustainability assessment of protein-soil composite materials for limited resource environments},
  author={Roedel, Henning and Plata, Isamar Rosa and Lepech, Michael and Loftus, David},
  journal={Journal of Renewable Materials},
  volume={3},
  number={3},
  pages={183--194},
  year={2015},
  publisher={Tech Science Press}
}

@inproceedings{bsc_4,
  title={Multiscale modeling and testing of protein-bound regolith and soils},
  author={Rosa, I and Lepech, MD and Loftus, DJ},
  booktitle={16th Biennial International Conference on Engineering, Science, Construction, and Operations in Challenging Environments},
  pages={580--590},
  year={2018},
  organization={American Society of Civil Engineers Reston, VA}
}

@article{bsc_5,
  title={Green recyclable biocomposite prepared from lignin and bamboo},
  author={Ren, Yi and Zhong, Yanan and Yang, Yang and Huo, Hongfeng and Zhang, Lei and Zhang, Jijuan and Huang, Kai and Zhang, Zhongfeng},
  journal={Journal of Cleaner Production},
  volume={449},
  pages={141710},
  year={2024},
  publisher={Elsevier}
}

@article{bsc_6,
  title={Life cycle assessment and design of LignoBlock: A lignin bound block on the path towards a green transition of the construction industry},
  author={Miao, Barney H and Headrick, Robert J and Li, Zhiye and Spanu, Leonardo and Loftus, David J and Lepech, Michael D},
  journal={Journal of Cleaner Production},
  volume={474},
  pages={143610},
  year={2024},
  publisher={Elsevier}
}

@article{bsc_7,
  title={Recycling of lignin-based biocomposites: Improving sustainability and enhancing material strength},
  author={Miao, Barney H and Woo, Daniel and Javan, Darius and Garboczi, Edward J and Headrick, Robert J and Lesh, Andrew C and Li, Zhiye and Loftus, David J and Lepech, Michael D},
  journal={Resources, Conservation and Recycling},
  volume={215},
  pages={108104},
  year={2025},
  publisher={Elsevier}
}

@inproceedings{bsc_8,
  title={Taking advantage of biological binders to solidify granular material: manufacture and recyclability of lignin-based biopolymer composites},
  author={Miao, Barney H and Lesh, Andrew C and Loftus, David J and Lepech, Michael D},
  booktitle={Biologically Inspired Materials, Processes, and Systems (BIMPS) 2025},
  volume={13430},
  pages={81--94},
  year={2025},
  organization={SPIE}
}

@article{bsc_9,
  title={Solvent selection enables sustainable and affordable lignin biocomposite for cement-free construction},
  author={Miao, Barney H and Woo, Daniel and Lesh, Andrew C and Loftus, David J and Lepech, Michael D},
  journal={Cleaner Materials},
  pages={100371},
  year={2026},
  publisher={Elsevier}
}

@article{bsc_11,
  title={Hypervelocity impact performance of biopolymer-bound soil composites for space construction},
  author={Allende, Maria I and Davis, B Alan and Miller, Joshua E and Christiansen, Eric L and Lepech, Michael D and Loftus, David J},
  journal={Journal of Aerospace Engineering},
  volume={33},
  number={2},
  pages={04020001},
  year={2020},
  publisher={American Society of Civil Engineers}
}

@article{bsc_12,
  title={Determining the yield stress of a biopolymer-bound soil composite for extrusion-based 3D printing applications},
  author={Biggerstaff, Adrian and Fuller, Gerald and Lepech, Michael and Loftus, David},
  journal={Construction and Building Materials},
  volume={305},
  pages={124730},
  year={2021},
  publisher={Elsevier}
}

@article{bsc_13,
  title={A shape stability model for 3D printable biopolymer-bound soil composite},
  author={Biggerstaff, Adrian and Lepech, Michael and Fuller, Gerald and Loftus, David},
  journal={Construction and Building Materials},
  volume={321},
  pages={126337},
  year={2022},
  publisher={Elsevier}
}

@article{bsc_14,
  title={Determining the structuration of biopolymer-bound soil composite},
  author={Biggerstaff, Adrian and Lepech, Michael and Loftus, David},
  journal={Materials and Structures},
  volume={55},
  number={7},
  pages={190},
  year={2022},
  publisher={Springer}
}

@article{bsc_15,
  title={A Study on the Flow Behavior and Thixotropy of Biopolymer-Bound Soil Composite},
  author={Biggerstaff, Adrian O and Lepech, Michael and Loftus, David},
  journal={Journal of Materials in Civil Engineering},
  volume={37},
  number={2},
  pages={04024487},
  year={2025},
  publisher={American Society of Civil Engineers}
}

@inproceedings{bsc_16,
  title={Biosys: efficient quality control system for manufacturing of sustainable biopolymer composites},
  author={Miao, Barney H and Dong, Yiwen and Theissler, Andreas and Lesh, Andrew C and Loftus, David J and Lepech, Michael D},
  booktitle={Proceedings of the 11th ACM International Conference on Systems for Energy-Efficient Buildings, Cities, and Transportation},
  pages={11--21},
  year={2024}
}

@article{bsc_17,
  title={AI-powered non-destructive testing for smart manufacturing of carbon-negative biopolymer-bound soil composite},
  author={Miao, Barney H and Dong, Yiwen and Theissler, Andreas and Lesh, Andrew C and Loftus, David J and Lepech, Michael D},
  journal={Communications Engineering},
  year={2026},
  publisher={Nature Publishing Group UK London}
}

@article{bsc_18,
  title={Prediction of ultimate compressive strength for biopolymer-bound soil composites (BSC) using sliding wingtip crack analysis},
  author={Roedel, Henning and Rosa, Isamar and Allende, Maria I and Lepech, Michael D and Loftus, David J and Garboczi, Edward J},
  journal={Engineering Fracture Mechanics},
  volume={218},
  pages={106570},
  year={2019},
  publisher={Elsevier}
}

@inproceedings{bsc_19,
  title={Creation of statistically equivalent periodic unit cells for protein-bound soils},
  author={Rosa, Isamar and Roedel, Henning and Lepech, Michael D and Loftus, David J},
  booktitle={ASME International Mechanical Engineering Congress and Exposition},
  volume={57526},
  pages={V009T12A058},
  year={2015},
  organization={American Society of Mechanical Engineers}
}

@article{bsc_20,
  title={A comprehensive review of various biopolymer composites and their applications: from biocompatibility to self-healing},
  author={Kumar, A and Mishra, RK and Verma, K and Aldosari, SM and Maity, CK and Verma, S and Patel, R and Thakur, VK},
  journal={Materials Today Sustainability},
  volume={23},
  pages={100431},
  year={2023},
  publisher={Elsevier}
}

@article{bsc_21,
  title={Synthetic biology for fibers, adhesives, and active camouflage materials in protection and aerospace},
  author={Roberts, Aled D and Finnigan, William and Wolde-Michael, Emmanuel and Kelly, Paul and Blaker, Jonny J and Hay, Sam and Breitling, Rainer and Takano, Eriko and Scrutton, Nigel S},
  journal={MRS communications},
  volume={9},
  number={2},
  pages={486--504},
  year={2019},
  publisher={Cambridge University Press}
}

@article{bsc_22,
  title={Blood, sweat, and tears: extraterrestrial regolith biocomposites with in vivo binders},
  author={Roberts, Ales Deakin and Whittall, DR and Breitling, Rainer and Takano, Eriko and Blaker, Jonny J and Hay, Sam and Scrutton, Nigel S},
  journal={Materials Today Bio},
  volume={12},
  pages={100136},
  year={2021},
  publisher={Elsevier}
}

@article{bsc_23,
  title={Effects of Xanthan gum biopolymer on soil strengthening},
  author={Chang, Ilhan and Im, Jooyoung and Prasidhi, Awlia Kharis and Cho, Gye-Chun},
  journal={Construction and Building Materials},
  volume={74},
  pages={65--72},
  year={2015},
  publisher={Elsevier}
}

@article{bsc_24,
  title={Evaluating the physical characteristics of biopolymer/soil mixtures},
  author={Ayeldeen, Mohamed K and Negm, Abdelazim M and El Sawwaf, Mostafa A},
  journal={Arabian Journal of Geosciences},
  volume={9},
  number={5},
  pages={371},
  year={2016},
  publisher={Springer}
}

@article{bsc_25,
  title={Geotechnical behavior of a beta-1, 3/1, 6-glucan biopolymer-treated residual soil},
  author={Chang, Ilhan and Cho, Gye-Chun},
  journal={Geomech. Eng},
  volume={7},
  number={6},
  pages={633--647},
  year={2014}
}

@article{bsc_26,
  title={Application of microbial biopolymers as an alternative construction binder for earth buildings in underdeveloped countries},
  author={Chang, Ilhan and Jeon, Minkyung and Cho, Gye-Chun},
  journal={International journal of polymer science},
  volume={2015},
  number={1},
  pages={326745},
  year={2015},
  publisher={Wiley Online Library}
}

@article{bsc_27,
  title={Improving mechanical properties of sand using biopolymers},
  author={Khatami, Hamid Reza and O’Kelly, Brendan C},
  journal={Journal of Geotechnical and Geoenvironmental Engineering},
  volume={139},
  number={8},
  pages={1402--1406},
  year={2013},
  publisher={American Society of Civil Engineers}
}

@article{bsc_28,
  title={Mechanical properties of biopolymer-stabilised soil-based construction materials},
  author={Muguda, Sravan and Booth, Samuel John and Hughes, Paul Neil and Augarde, Charles Edward and Perlot, Celine and Bruno, Agostino Walter and Gallipoli, Domenico},
  journal={G{\'e}otechnique letters},
  volume={7},
  number={4},
  pages={309--314},
  year={2017},
  publisher={Thomas Telford Ltd}
}

@article{bsc_29,
  title={Seaweed biopolymers as additives for unfired clay bricks},
  author={Dove, Cassandra A and Bradley, Fiona F and Patwardhan, Siddharth V},
  journal={Materials and Structures},
  volume={49},
  number={11},
  pages={4463--4482},
  year={2016},
  publisher={Springer}
}

@article{bsc_30,
  title={StarCrete: A starch-based biocomposite for off-world construction},
  author={Roberts, Aled D and Scrutton, Nigel S},
  journal={Open Engineering},
  volume={13},
  number={1},
  year={2023},
  publisher={De Gruyter Open Access}
}

\clearpage
\appendix

\label{sec:appendix}
\section{Appendix}

\begin{table*}[!h]
\centering
\small
\setlength{\tabcolsep}{3pt}
\begin{tabular}{lcccccccccc}
\toprule
& \multicolumn{2}{c}{Gemma2}
& \multicolumn{2}{c}{Llama3}
& \multicolumn{2}{c}{Llama3.1}
& \multicolumn{2}{c}{Mistral}
& \multicolumn{2}{c}{Qwen2.5} \\
\cmidrule(lr){2-3} \cmidrule(lr){4-5} \cmidrule(lr){6-7} \cmidrule(lr){8-9} \cmidrule(lr){10-11}
& gpt-4o & gemini
& gpt-4o & gemini
& gpt-4o & gemini
& gpt-4o & gemini
& gpt-4o & gemini \\
\midrule
Baseline &22.36 &15.93 &17.28 & 12.73&18.73 & 17.45&16.20 &13.55 &22.36 &16.63 \\
Vanilla-RAG   & 69.40 & 72.06 & 56.56 & 62.40 & 88.70 & 82.56 & 58.22 & 77.68 & \textbf{89.00} & 85.80 \\
EGT-KG(AS)  & 74.58 & 78.18 & 75.96 & 74.70 & 89.72 & 84.62 & 57.30 & \textbf{79.00} & 83.86 & 86.21 \\
EGT-KG(ES) & \textbf{77.36} & \textbf{79.98} & \textbf{78.72} & \textbf{78.00} & \textbf{89.78} & \textbf{86.52} & \textbf{58.88} & 75.90 & 88.42 & \textbf{87.00} \\
\bottomrule
\end{tabular}

\vspace{0.5\baselineskip}
\caption{Performance comparison across answer models and judge models for simple questions ($question_1$)}
\label{tab:simple_results}
\vspace{0.5\baselineskip}
\end{table*}

\begin{table*}[!t]
\centering
\small
\setlength{\tabcolsep}{3pt}
\begin{tabular}{lcccccccccc}
\toprule
& \multicolumn{2}{c}{Gemma2}
& \multicolumn{2}{c}{Llama3}
& \multicolumn{2}{c}{Llama3.1}
& \multicolumn{2}{c}{Mistral}
& \multicolumn{2}{c}{Qwen2.5} \\
\cmidrule(lr){2-3} \cmidrule(lr){4-5} \cmidrule(lr){6-7} \cmidrule(lr){8-9} \cmidrule(lr){10-11}
& gpt-4o & gemini
& gpt-4o & gemini
& gpt-4o & gemini
& gpt-4o & gemini
& gpt-4o & gemini \\
\midrule
Baseline & 49.55& 33.28& 38.76& 36.88&39.65 &32.45 &41.10 & 31.03&48.02 & 32.23 \\
Vanilla-RAG   & 64.12 & \textbf{69.10} & \textbf{66.18} & \textbf{62.48} & 79.90 & 72.10 & 60.42 & \textbf{77.42} & \textbf{83.02} & 70.38 \\
EGT-KG(AS)  & \textbf{69.86} & 67.06 & 64.78 & 57.52 & 77.78 & 72.26 & \textbf{63.38} & 76.56 & 81.78 & \textbf{74.98} \\
EGT-KG(ES) & 68.88 & 68.10 & 58.92 & 58.42 & \textbf{82.26} & \textbf{76.92} & 62.26 & 72.56 & 81.98 & 72.82 \\
\bottomrule
\end{tabular}

\vspace{0.5\baselineskip}
\caption{Performance comparison across answer models and judge models for complex questions ($question_2$)}
\label{tab:complex_results}
\vspace{0.5\baselineskip}
\end{table*}

\subsection{Additional performance analysis}
\label{app:additional}

Table~\ref{tab:simple_results} and ~\ref{tab:complex_results} together show the workflow performance for simple questions and complex questions. For simple questions, both EGT-KG variants outperform the vanilla RAG workflow in 9 out of 10 comparisons, with ES EGT-KG variants achieving the best result in 8 of them. However, the benefit becomes much less stable on complex questions, suggesting that our framework primarily improves precision in evidence localization for simple fact retrieval questions. For complex questions with multi-hop reasoning and global evidence aggregation, the benefits from finer-grained schemas are reduced. This finding further confirms the limitation discussed before: the one-hop query expansion reduces model's capability in processing complex queries that require multi-hop reasoning or cross-document inference. Notably, the baseline workflow (plain QA with SLMs) achieves relatively higher Final Scores on complex questions than on simple factual questions. A plausible explanation is that the complex questions in our benchmark are broader and more discussion-oriented. Thus, even without external retrieval, SLMs can still obtain partial credit on complex questions by generating generally relevant responses. Table~\ref{tab:dimension_results} reports the raw scores for the six S3CRF dimensions. As expected, the gains in soundness (+3.03\%), correctness (+4.52\%), completeness (+2.78\%), and relevance (+3.20\%) are noticeably larger than conciseness (+2.22\%) and fluency (+1.71\%). It suggests that the main benefit of our framework comes from improving core answer-quality dimensions related to reasoning, factual accuracy, coverage, and query alignment, rather than surface-level language style. 

\begin{table*}[!h]
\centering
\small
\setlength{\tabcolsep}{3pt}
\begin{tabular}{lcccccccccccc}
\toprule
& \multicolumn{2}{c}{Soundness}
& \multicolumn{2}{c}{Correctness}
& \multicolumn{2}{c}{Completeness}
& \multicolumn{2}{c}{Conciseness}
& \multicolumn{2}{c}{Relevance}
& \multicolumn{2}{c}{Fluency} \\
\cmidrule(lr){2-3} \cmidrule(lr){4-5} \cmidrule(lr){6-7} \cmidrule(lr){8-9} \cmidrule(lr){10-11} \cmidrule(lr){12-13}
& gpt & gemini
& gpt & gemini
& gpt & gemini
& gpt & gemini
& gpt & gemini
& gpt & gemini \\
\midrule
Baseline    & 29.0 & 18.6 & 19.4 & 15.1 & 29.0 & 18.6 & 60.1 & 58.3 & 33.1 & 23.9 & 75.8 & 89.2 \\
Vanilla-RAG   & 70.3 & 71.5 & 68.8 & 69.7 & 60.3 & 50.8 & 83.8 & 91.8 & 78.2 & 86.6 & 88.5 & \textbf{96.7} \\
EGT-KG(AS)  & 71.8 & \textbf{74.1} & 72.2 & 71.2 & \textbf{61.4} & 51.8 & 86.7 & \textbf{92.9} & 81.1 & 89.5 & 90.6 & 96.5 \\
EGT-KG(ES) & \textbf{72.4} & 73.9 & \textbf{73.9} & \textbf{72.2} & \textbf{61.4} & \textbf{53.6} & \textbf{86.8} & 91.0 & \textbf{82.0} & \textbf{90.1} & \textbf{91.2} & \textbf{96.7} \\
\bottomrule
\end{tabular}

\vspace{0.5\baselineskip}
\caption{Dimensional performance comparison across workflows.}
\label{tab:dimension_results}
\vspace{0.5\baselineskip}
\end{table*}

\begin{table*}[!t]
\centering
\small
\begin{tabular}{lrrrrrr}
\toprule
\textbf{Setting} & \textbf{gemma2} & \textbf{qwen2.5} & \textbf{llama3.1} & \textbf{llama3} & \textbf{mistral} & \textbf{Mean} \\
\midrule
Q-only baseline & 0.1338 & 0.1370 & 0.1471 & 0.1231 & 0.1253 & 0.1333 \\
Vanilla RAG & 0.1432 & 0.1590 & 0.1729 & \textbf{0.1737} & 0.1449 & 0.1587 \\
EGT-KG & \textbf{0.1554} & \textbf{0.1641} & \textbf{0.1840} & 0.1710 & \textbf{0.1659} & \textbf{0.1681} \\
\midrule
EGT-KG vs. Vanilla & +8.5\% & +3.2\% & +6.4\% & -1.6\% & +14.5\% & +5.9\% \\
\bottomrule
\end{tabular}
\caption{Supplementary QASPER results. Answer F1 is reported over the full development split of 1,005 questions.}
\label{tab:qasper_crosspaper}
\end{table*}

\subsection{Component ablation}

\begin{table*}[!h]
\centering
\small
\begin{tabular}{lccccc}
\toprule
\textbf{Judge} & \textbf{A} & \textbf{B} & \textbf{C} & \textbf{D} & \textbf{E} \\
 & vanilla dense & A + query expansion & B\,+\,rerank & B\,+\,window & full \\
\midrule
\textit{gpt-4o}         & 72.01 & 75.28 \small{(+3.27)} & 74.62 \small{(-0.66)} & 75.25 \small{(-0.03)} & \textbf{76.30} \small{(+4.29)} \\
\textit{gemini-2.5-pro} & 66.60 & 68.38 \small{(+1.78)} & 70.41 \small{(+2.03)} & 68.40 \small{(+0.02)} & \textbf{70.46} \small{(+3.86)} \\
\bottomrule
\end{tabular}
\caption{Component ablation, pooled Final Score over the five answer models
(40 questions $\times$ 5 sampling rounds, $n=200$ per cell).
A is vanilla dense retrieval ($\alpha=1.0$, no re-ranking, no evidence window);
B adds knowledge-graph query expansion ($\alpha=0.7$);
C and D add evidence re-ranking and the evidence window to B, respectively.
Parentheses give the increment over A for B and E, and over B for C and D.}
\label{tab:ablation}
\end{table*}

\begin{table*}[h!]
\centering
\small
\begin{tabular}{lccc}
\toprule
Retrieval pool & Vanilla RAG & EGT-KG & $\Delta$ \\
\midrule
Official distractor (10) & 0.4857 & 0.5015 & +0.016 \\
Pooled corpus (1,977)    & 0.3165 & 0.4585 & \textbf{+0.142} \\
\bottomrule
\end{tabular}
\caption{HotpotQA Answer F1 under two retrieval regimes, 200 development
questions, \textit{gemma2:9b}.}
\label{tab:hotpot}
\end{table*}

As Table \ref{tab:ablation} shows, no single component makes a dominant contribution to the overall improvement, because its measured effect flips between the judge models. To be specific, the re-ranker carries the effect under \textit{gemini-2.5-pro} (+2.03 over query expansion), whereas the same addition is negative under \textit{gpt-4o} (-0.66), and the evidence window on its own contributes essentially nothing over query expansion under either judge (+0.02 and -0.03). The stable partial gains is C (query expansion + rerank): because evidence re-ranking operates on the candidate chunks that query expansion supplies, the two components together consistently improve over vanilla RAG under both judges (+2.61 from \textit{gpt-4o}, +3.81 from \textit{gemini-2.5-pro}), even though the marginal contribution of re-ranking alone changes sign. The assembled pipeline is therefore the best configuration to deploy. The full framework is the highest-scoring configuration under both judges (A $\to$ E: +4.29 from \textit{gpt-4o}; +3.86 from \textit{gemini-2.5-pro}), and no partial configuration beats it. The remaining improvements come from the interaction between evidence re-ranking and the evidence window. The mechanism is straightforward: The evidence window can suppress noise and highlight the most relevant sentence. But if only the evidence window is deployed, when the gold evidence is not present in the retrieved candidate chunks, the answer model is still stuck for lack of ground truth — which is exactly why the evidence window alone changes nothing. Once re-ranking has moved the chunk containing the gold evidence into the final top-k, the evidence window pays off: it removes the unnecessary surrounding text, cutting mean context by 22\% ($8,196 \to 6,399$ characters) at no cost in score, and lets the answer model focus on the evidence that matters.

\subsection{Supplementary QASPER Evaluation}
\label{app:qasper}

To further study whether the EGT-KG framework generalizes beyond the BSC corpus, we conduct a supplementary evaluation on QASPER~\citep{qasper}. QASPER is a scientific question-answering benchmark built from NLP research papers. Each question is associated with a source paper and is answered using evidence from that paper.

The official QASPER evaluation protocol provides the target paper ID and restricts retrieval to chunks from that paper. This per-paper setting is appropriate for evaluating reading comprehension once the relevant paper is known, but it does not fully reflect many practical scientific RAG scenarios, where a user asks a question without specifying which paper contains the answer. We therefore use a cross-paper retrieval setting: the system is not given the gold paper ID and must retrieve evidence from all papers in the development split. In our setup, retrieval is performed over 5,253 chunks from 281 papers, and evaluation is conducted on all 1,005 development questions.

We compare three settings. The first is a question-only baseline, where the models answer questions without retrieved context. The second is vanilla cross-paper RAG, which retrieves chunks from the full cross-paper pool using dense retrieval. The third is our EGT-KG framework, which augments retrieval with KG-expanded query terms, evidence-aware reranking, and evidence-window construction. We report Answer F1 using the official QASPER evaluator.

The results show that retrieval is essential in this more realistic cross-paper setting, where vanilla RAG improves mean Answer F1 from 0.1333 to 0.1587 (19.0\%) over the question-only baseline. EGT-KG further improves mean Answer F1 to 0.1681, a 5.9\% relative gain over vanilla RAG, with positive gains for four out of five tested models. 
Additionally, we performed a simple test on official per-paper setting on \textit{gemma2:9b} ($\alpha = 0.9,\lambda=0.5,k=8$). The EGT-KG framework (0.367) outperform vanilla RAG (0.334) by 10.2\%. The improvement on scores are derived from more relevant chunk candidates for both workflows.

\subsection{Cross-domain evaluation on HotpotQA}
\label{app:hotpot}

We evaluated on 200 HotpotQA development questions for generalization,
inducing the relation schema from that corpus's own relation-label distribution.

The outcome depends entirely on how large a candidate pool the retrieval has to
search. In the official distractor setting, each question is supplied with ten
candidate paragraphs with two gold paragraphs, and the framework adds nothing
($+0.016$): dense retrieval already finds the gold paragraphs, so there
is no localization problem left to solve. Pooling the same questions' paragraphs
into a single 1,977-paragraph corpus reverses this ($+0.142$).
Growing the pool from 10 to 1,977 candidates costs vanilla RAG 34.8\% of its
Answer F1 and EGT-KG 8.6\%, and the complete gold-paragraph set is retrieved for
36.0\% of questions by vanilla RAG against 56.0\% by EGT-KG.

\subsection{Small-corpus construction setting}

Another practical advantage of our framework is the feasibility of being built on a small corpus in a limited resource scenario. Rather than requiring large-scale indexing or computationally intensive global graph structure, our EGT-KG framework constructs a knowledge graph from a compact benchmark of 30 papers. This design decision mimics the real-world situation in the emerging scientific domain of question-answering, where users may only have access to a small collection of relevant papers and may not be willing to afford the cost of heavy knowledge graph construction. Thus, our contribution to EGT-KG is to improve scientific QA while preserving the possibility of practical deployment. The experiment results show that even the lightweight knowledge graph, which is further trimmed by type schema, can provide useful and explainable retrieval to enhance response generation for scientific QA.

\subsection{Introduction to BSC}
\label{app:bsc}
Biopolymer-bound soil composite (BSC) is a nature-inspired material that valorizes waste-stream biopolymers from major industries (e.g., paper pulping, biofuels, and agriculture) as binders for granular materials. Unlike conventional materials such as ordinary Portland cement concrete, BSC is potentially carbon-negative due to its use of carbon-rich biopolymers, enabling sequestration of carbon in the composite \citep{miao2024life}. It is also fully recyclable, reduces construction waste, and offers compressive strength comparable to concrete, making it a promising low-cost, sustainable, cement-free alternative for the built environment.
\clearpage
\onecolumn

\subsection{Detailed Pipeline Diagrams}
\label{app:pipeline_details}

\begin{center}
\includegraphics[width=\textwidth,height=0.30\textheight,keepaspectratio]{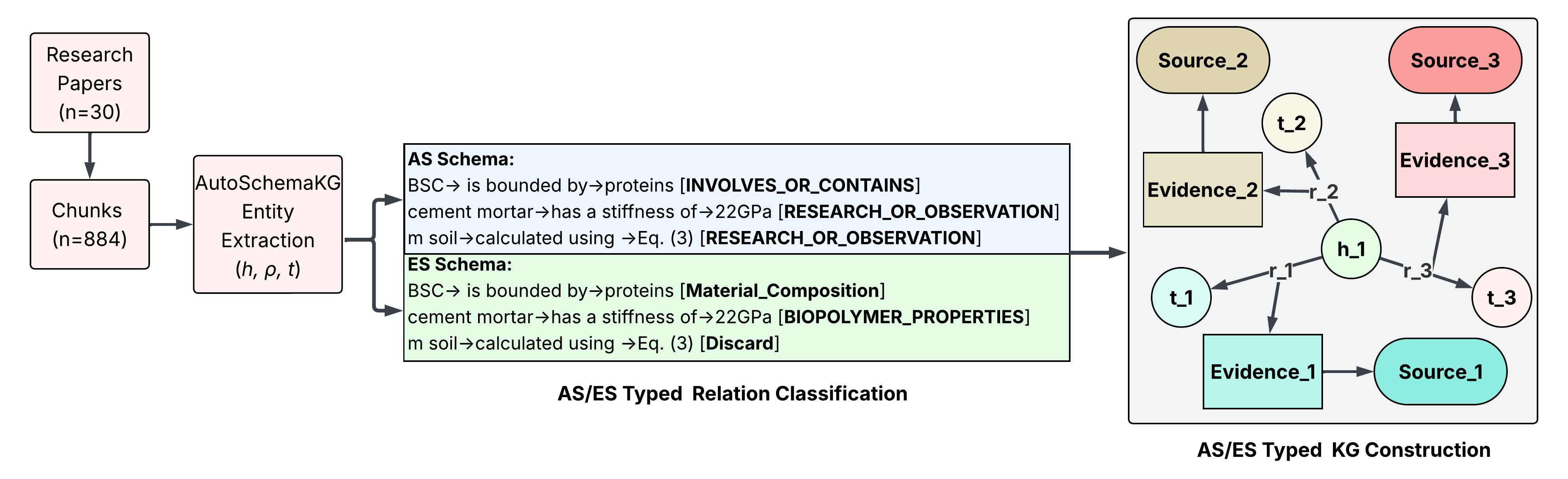}

\vspace{0.3\baselineskip}
\captionof{figure}{Stage 1 of EGT-KG: triples extraction, relation classification, and KG construction.}
\label{kg-pipe}

\vspace{1.0em}

\includegraphics[width=\textwidth,height=0.34\textheight,keepaspectratio]{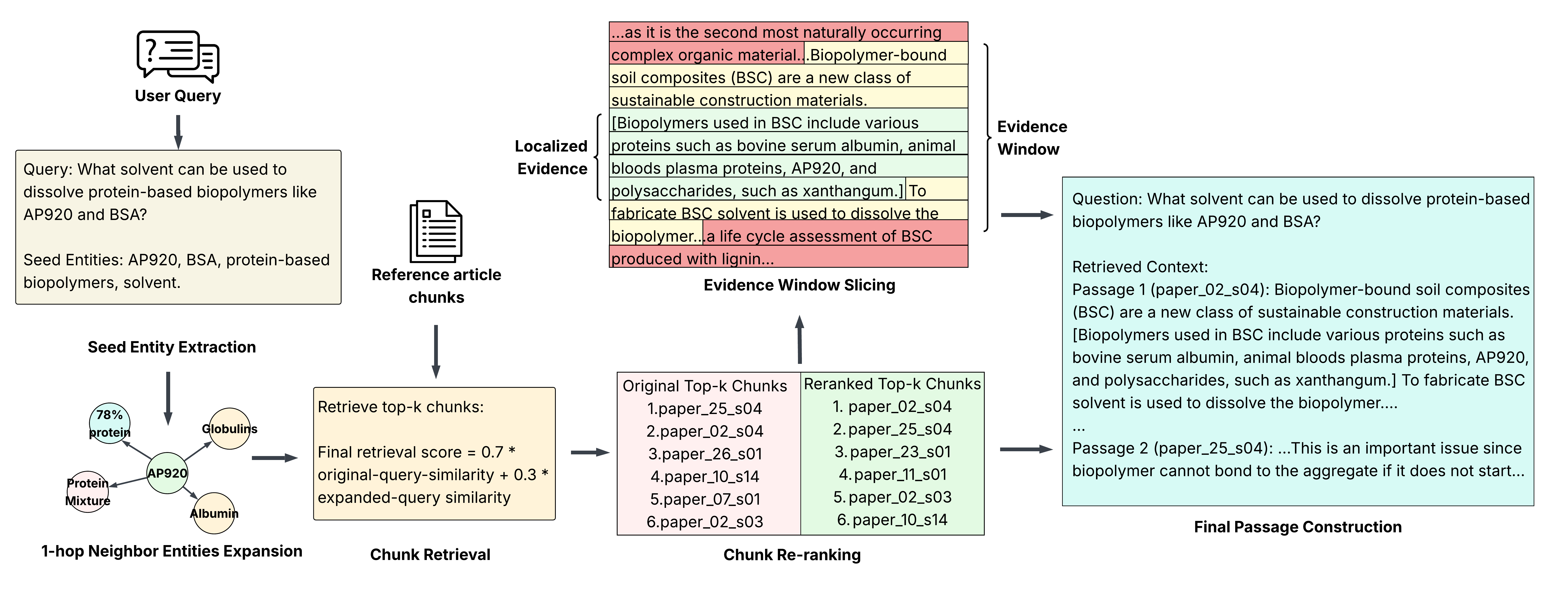}

\vspace{0.3\baselineskip}
\captionof{figure}{Stage 2 of EGT-KG: query expansion, context retrieval and re-ranking, evidence-aware context slicing, and passage construction.}
\label{stage2}
\end{center}

\clearpage

\subsection{Prompts}
\label{app:prompts}

\begin{center}
\includegraphics[width=\textwidth,height=0.36\textheight,keepaspectratio]{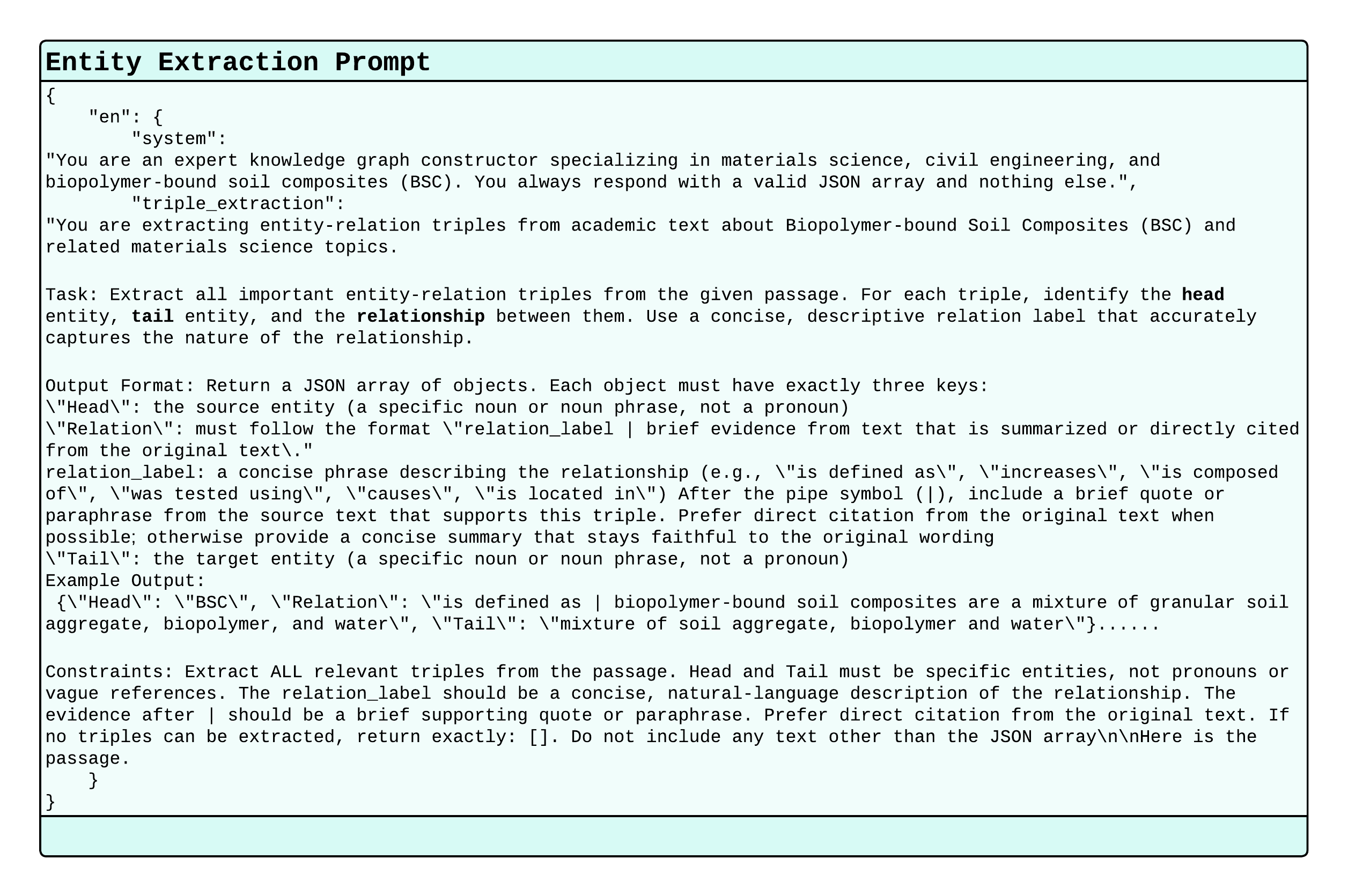}

\vspace{0.3\baselineskip}
\captionof{figure}{Entity extraction prompt}
\label{p_1}

\vspace{1.0em}

\includegraphics[width=\textwidth,height=0.28\textheight,keepaspectratio]{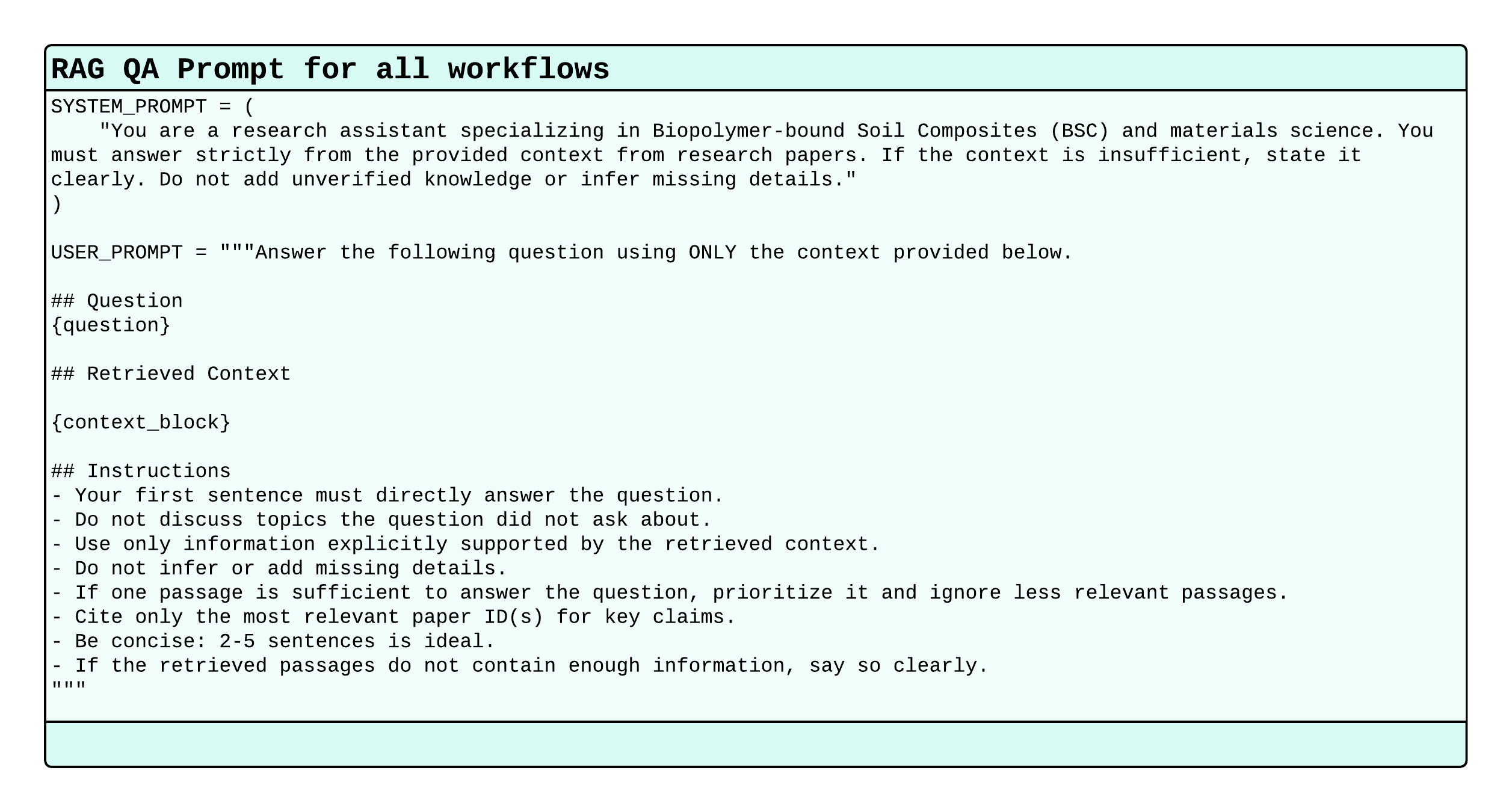}

\vspace{0.3\baselineskip}
\captionof{figure}{RAG QA prompt for all workflows, containing original question, retrieved context, and instructions}
\label{p_rag_qa}

\vspace{1.0em}

\includegraphics[width=\textwidth,height=0.28\textheight,keepaspectratio]{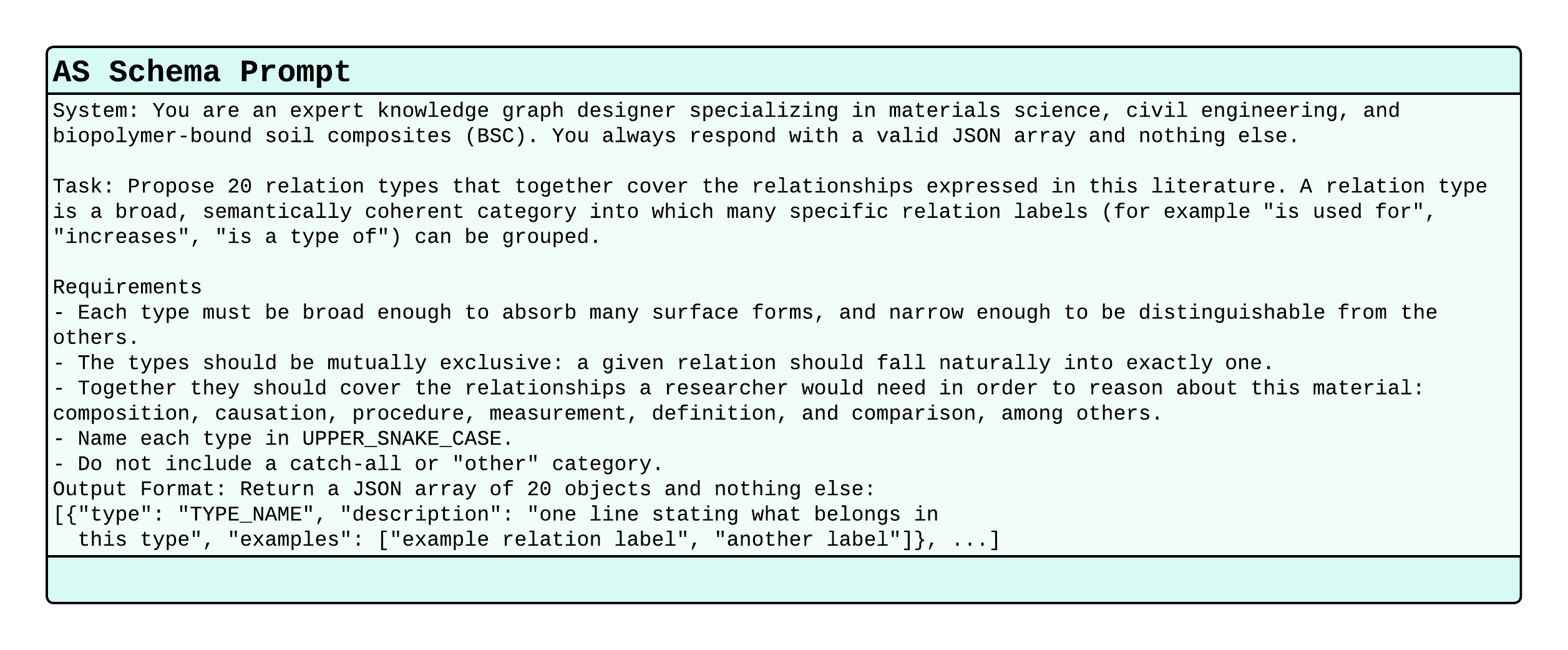}

\vspace{0.3\baselineskip}
\captionof{figure}{Prompt used to induce the AS relation schema.}
\label{as_schema}

\end{center}

\subsection{Parameter sensitivity}
\label{app:params}

\begin{table}[h]
\centering
\small
\begin{tabular}{lcccc}
\toprule
$\lambda$ & 0.40 & 0.50 & 0.65 & 0.80 \\
\midrule
Final Score & 66.47 & 68.37 & \textbf{73.54} & 69.12 \\
\midrule
$k$ & 3 & 6 & 10 & 15 \\
\midrule
Final Score & 63.80 & \textbf{71.40} & 68.20 & 66.30 \\
\bottomrule
\end{tabular}
\caption{Sensitivity to the re-ranking weight $\lambda$ and to the number of
retrieved chunks $k$. Reported results use $\lambda=0.65$ and $k=6$.}
\label{tab:params}
\end{table}

\subsection{Performance by relation type}
\label{app:reltype}

\begin{table}[h]
\centering
\small
\begin{tabular}{lrrrr}
\toprule
Relation type & KG\% & Ret\% & ratio & $\Delta$ \\
\midrule
Causes or Influences     & 36.4 & 34.5 & 0.95 & +3.96 \\
Research Observations    & 17.1 & 16.2 & 0.95 & +4.98 \\
Material Composition     & 11.9 & 12.0 & 1.01 & +2.03 \\
Manufacture Procedure    & 10.3 &  9.8 & 0.95 & +4.72 \\
Logical Reasoning        &  7.5 &  7.7 & 1.03 & +2.83 \\
Definition               &  5.7 &  8.4 & 1.47 & +1.70 \\
Experimental Methods     &  5.0 &  5.8 & 1.16 & +5.03 \\
Biopolymer Properties    &  3.8 &  2.4 & 0.63 & +5.49 \\
Temporal Relation        &  1.0 &  1.2 & 1.20 & +5.59 \\
Design Procedure         &  0.9 &  1.5 & 1.67 & $-$0.20 \\
Experimental Procedures  &  0.4 &  0.5 & 1.25 & +2.78 \\
\bottomrule
\end{tabular}
\caption{Performance by relation type under the ES schema. KG\% is the type's
share of the graph, Ret\% is the share of the evidence actually retrieved. $\Delta$ is the Final Score improvement over
vanilla RAG.}
\label{tab:reltype}
\end{table}

\subsection{Distribution of relation types in the reified knowledge graph}

\begin{center}
\includegraphics[width=0.9\textwidth,height=0.38\textheight,keepaspectratio]{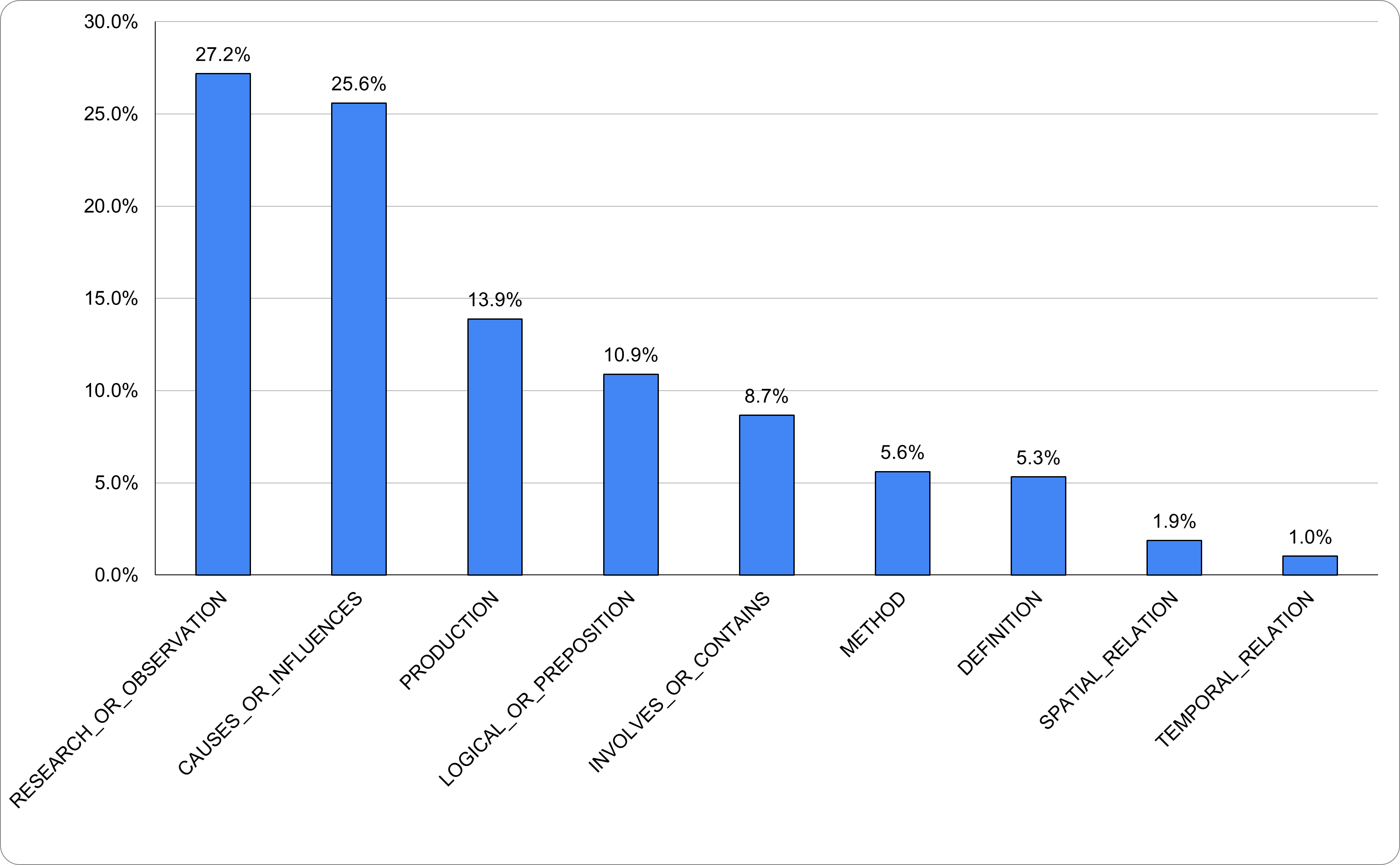}

\vspace{0.4\baselineskip}
\captionof{figure}{EGT-KG(AS) relation schema distribution}
\label{9-type-distribution}

\vspace{1.2em}

\includegraphics[width=0.9\textwidth,height=0.38\textheight,keepaspectratio]{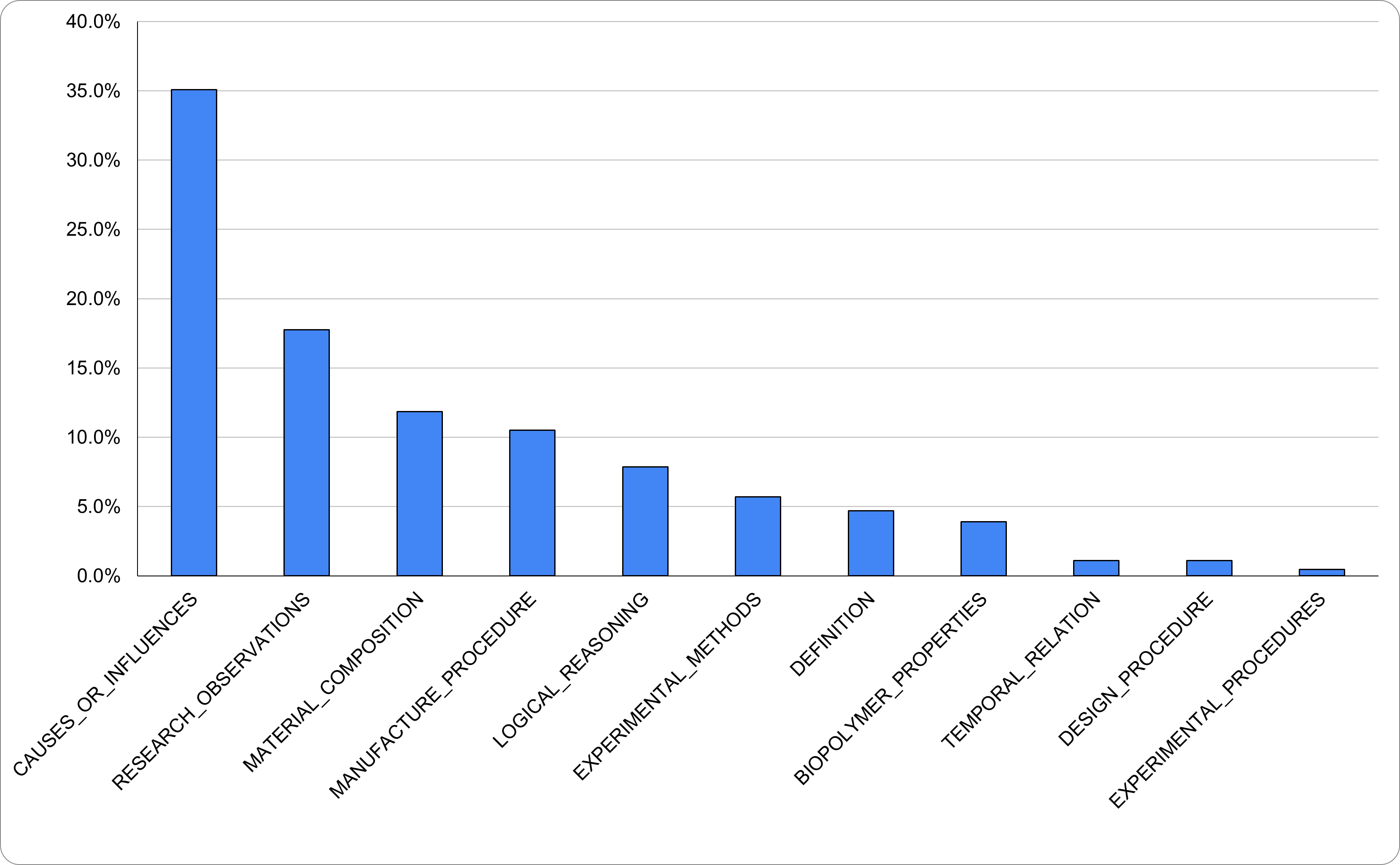}

\vspace{0.4\baselineskip}
\captionof{figure}{EGT-KG(ES) relation schema distribution}
\label{11-type-distribution}
\end{center}

\clearpage
\onecolumn

\subsection{Evaluation Criteria}

\begin{center}
\small
\begin{tabularx}{\textwidth}{p{0.16\textwidth} X p{0.18\textwidth}}
\toprule
\textbf{Metric} & \textbf{Definition} & \textbf{Primary focus} \\
\midrule
Soundness & Logical validity of reasoning from question to answer, including appropriate use of referenced material. & Reasoning validity \\
Correctness & Factual accuracy compared to the reference material; hallucinated facts reduce the score. & Factual grounding \\
Completeness & Coverage of all relevant aspects of the question. & Information coverage \\
Conciseness & Appropriate detail and brevity without unnecessary verbosity. & Brevity and clarity \\
Relevance & Alignment with the user's query. & Query alignment \\
Fluency & Grammatical, coherent, and readable language. & Readability \\
\bottomrule
\end{tabularx}
\captionof{table}{S3CRF evaluation metrics.}
\label{tab:S3CRF}
\end{center}

\begin{center}
\small
\setlength{\tabcolsep}{5pt}
\renewcommand{\arraystretch}{1.12}
\begin{tabularx}{\textwidth}{>{\raggedright\arraybackslash}p{0.17\textwidth} c >{\raggedright\arraybackslash}X}
\toprule
\textbf{Dimension} & \textbf{Score} & \textbf{Description} \\
\midrule

\multirow{6}{0.17\textwidth}{\raggedright\textbf{Soundness} \\ $w_{snd}=0.275$}
& 0 & Incorrect logical connection between question and answer. \\
& 1 & Poor reasoning, conclusion is not directly supported. \\
& 2 & Weak reasoning, several logical flaws in answer. \\
& 3 & Acceptable reasoning with partially unsupported claims. \\
& 4 & Strong reasoning with minor gaps. \\
& 5 & Perfect logical reasoning with proper use of evidence. \\
\addlinespace[0.4em]
\hline
\multirow{6}{0.17\textwidth}{\raggedright\textbf{Correctness} \\$w_{cor}=0.275$}
& 0 & Incorrect or fabricated response. \\
& 1 & Significant factual errors or hallucinations. \\
& 2 & Multiple minor factual errors or hallucinations. \\
& 3 & Generally correct with acceptable inaccuracies. \\
& 4 & Mostly accurate with trivial errors. \\
& 5 & Accurate and verifiable answer supported by reference documents. \\
\addlinespace[0.4em]
\hline
\multirow{6}{0.17\textwidth}{\raggedright\textbf{Completeness} \\$w_{cmp}=0.15$}
& 0 & Fails to address the question. \\
& 1 & Incomplete response with superficial coverage. \\
& 2 & Missing coverage of important information. \\
& 3 & Addresses main point but misses some details. \\
& 4 & Covers most important aspects. \\
& 5 & Comprehensively addresses all aspects. \\
\addlinespace[0.4em]
\hline
\multirow{6}{0.17\textwidth}{\raggedright\textbf{Conciseness}\\$w_{con}=0.05$}
& 0 & Completely unnecessary length. \\
& 1 & Severely verbose or oversimplified. \\
& 2 & Relatively too long/short, affecting readability. \\
& 3 & Somewhat verbose or oversimplified with acceptable readability. \\
& 4 & Mostly appropriate length with minor issues. \\
& 5 & Perfect balance of detail and brevity. \\
\addlinespace[0.4em]
\hline
\multirow{6}{0.17\textwidth}{\raggedright\textbf{Relevance}\\$w_{rel}=0.2$}
& 0 & Completely off-topic response. \\
& 1 & Mostly irrelevant with trivial relevant discussion. \\
& 2 & Partially relevant, significant off-topic content. \\
& 3 & Mostly relevant but some off-topic content. \\
& 4 & Highly relevant with minor off-topic content. \\
& 5 & Directly and precisely addresses the question. \\
\addlinespace[0.4em]
\hline
\multirow{6}{0.17\textwidth}{\raggedright\textbf{Fluency} \\$w_{flu}=0.05$}
& 0 & Incoherent. \\
& 1 & Poor organization, very hard to read. \\
& 2 & Choppy expression or hard to follow. \\
& 3 & Acceptable expression with some disjointed sections. \\
& 4 & Good coherence with minor awkwardness. \\
& 5 & Excellent flow, reads naturally. \\

\bottomrule
\end{tabularx}

\vspace{0.5\baselineskip}
\captionof{table}{Rubric for S3CRF evaluation (0-5).}
\label{rubric-vertical}
\end{center}

\clearpage

\subsection{Simple/complex questions in the question set and papers used in corpus}

\begin{center}
\small
\setlength{\tabcolsep}{5pt}
\renewcommand{\arraystretch}{1.05}
\begin{tabularx}{\textwidth}{p{0.06\textwidth} X}
\toprule
\textbf{ID} & \textbf{Question} \\
\midrule
Q1  & What solvent can be used to dissolve protein-based biopolymers like AP920 and BSA?\\
\hline
Q2  & Give me an overview of the processes needed to dissolve the biopolymers (e.g. time and any other considerations)\\
\hline
Q3  &What is the maximum amount (in percentage) of biopolymer that can be dissolved into solution?\\
\hline
Q4  & What parameter do we need to consider in order to maximize the amount of biopolymer in the composite?\\
\hline
Q5  & What type of compaction method is used to manufacture BSC? Why is this important?\\
\hline
Q6  & Why is BSC a sustainable building material? \\
\hline
Q7  & What happens with saturation ratios more than 1.0?\\
\hline
Q8  & How do we desiccate BSC?\\
\hline
Q9  & What is the porosity of BSC?\\
\hline
Q10 & How can I maximize the strength of these biocomposites?\\
\hline
Q11 & What biopolymers can be used to make BSC?\\
\hline
Q12 & What is the microstructure of BSC? \\
\hline
Q13 & How can the structure of BSC be modeled?\\
\hline
Q14 & How can I prevent shrinkage cracking in BSC?\\
\hline
Q15 & How can BSC structure be imaged?\\
\hline
Q16 & How can BSC be recycled?\\
\hline
Q17 & How is the strength of BSC characterized?\\
\hline
Q18 & What form of sample is used for BSC characterization?\\
\hline
Q19 & How can the sustainability of BSC be improved? \\
\hline
Q20 & How does the use of BSA contrast with AP920?\\
\bottomrule
\end{tabularx}

\vspace{0.5\baselineskip}
\captionof{table}{Simple question set used in QA evaluation.}
\label{tab:simple_questions}
\end{center}

\clearpage

\begin{center}
\small
\setlength{\tabcolsep}{5pt}
\renewcommand{\arraystretch}{1.05}
\begin{tabularx}{\textwidth}{p{0.06\textwidth} X}
\toprule
\textbf{ID} & \textbf{Question} \\
\midrule
Q1  & How can I select the most optimal solvent to dissolve a given biopolymer? How can I choose a solvent that not only effectively dissolves the biopolymer but also enhances the resulting bridge strength or interfacial bond strength between the aggregate and the biopolymer binder?\\
\hline
Q2  & How can each step in the manufacture of BSC be modified to improve both manufacturability and the overall macroscale properties of the composite? What potential issues could arise at each step, and how might they affect the final properties of the manufactured composite?\\
\hline
Q3  & Explain the relationship between biopolymer concentration and biopolymer solution viscosity during dissolution, and how the "workability limit" physically constrains the maximum binder content.\\
\hline
Q4  & If higher biopolymer content typically results in a stronger BSC, could we reduce the amount of compaction applied to the specimens (i.e., increase the void space in the material)? Would this reduction in compaction and the corresponding increase in void space, which the biopolymer can fill, affect any other material or design parameters?\\
\hline
Q5  & How can higher levels of compaction enhance the compressive strength of BSC, and what are the resulting impacts on design parameters?\\
\hline
Q6  & In comparison to hydrophilic BSCs which use water as a solvent, does BSC using hydrophobic biopolymers with a higher potential for carbon sequestration (they are more carbon rich) lead to always having a lower carbon footprint? \\
\hline
Q7  & If a novel biopolymer is being used and its particle density is not yet characterized, is it preferable to prepare an initial mix above or below the saturation ratio? What are the potential advantages and disadvantages of each approach, and which strategy would minimize risk in terms of workability, mechanical performance, and consistency of the resulting BSC?\\
\hline
Q8  & How do the desiccation requirements differ between protein-based BSCs (which use water as a solvent) and lignin-based BSCs (which use DMSO as a solvent)? If the desiccation time for a BSC using a particular solvent is excessively long, what strategies can be employed to accelerate the process without compromising the composite's mechanical integrity or interfacial bonding?\\
\hline
Q9  & Can microscale characterization techniques, such as X-ray micro-CT, be used to accurately determine the porosity of BSC? How can the porosity of BSC be increased without compromising its mechanical or physical properties?\\
\hline
Q10 & Rank the most effective methods for improving the compressive strength of biopolymer-bound soil composites, and provide a detailed explanation for each method. In your evaluation, consider not only the potential gains in compressive strength but also factors such as practicality, cost-effectiveness, scalability, and environmental footprint.\\
\hline
Q11 & Which biopolymers are most promising in terms of environmental impact, availability, and practicality. If I want to explore or develop a new variant of BSC, what are some promising biopolymers that can be used?\\
\hline
Q12 & What phenomena can be observed from microstructural images of BSC (phenomenon and type of image). \\
\hline
Q13 & What types of data are required to develop an accurate model of BSC behavior? Which material, structural, and processing parameters should be prioritized for measurement in future investigations to improve model reliability and predictive capability?\\
\hline
Q14 & Does a staggered desiccation approach, in which the BSC is first desiccated under ambient conditions and then oven-dried directly reduce the formation of shrinkage cracks? Please hypothesize how this potential approach can be designed (number of steps, time per step, conditions at each step)\\
\hline
Q15 & Can SEM be used in the same capacity/function as Micro-CT in calculating statistical descriptors (lineal path function and two-point probability correlation functions).\\
\hline
Q16 & Explain the process of recycling BSCs at the end of their lifecycle to reclaim the biopolymer. Also hypothesize a more effective way for recycling other than this approach.\\
\hline
Q17 & If I wanted to perform a study on the strength of a biopolymer bridge between two aggregate particles how would I go about this? Propose a plan.\\
\hline
Q18 & For macroscale considerations, are there any limits or standards for minimum size of a test specimen? What are these standards based on?\\
\hline
Q19 & Propose the most viable and potential studies that can be carried out. When proposing plans or studies, please consider how we can improve the affordability or accessibility of existing and future variants of BSC . \\
\hline
Q20 & Is BSA or AP920 more practical for construction in extraterrestrial environments, both in the short term and the long term?\\
\bottomrule
\end{tabularx}

\vspace{0.5\baselineskip}
\captionof{table}{Complex question set used in QA evaluation.}
\label{tab:complex_questions}
\end{center}

\clearpage

\begin{center}
\scriptsize
\setlength{\tabcolsep}{4pt}
\renewcommand{\arraystretch}{1.0}
\begin{tabularx}{\textwidth}{p{0.04\textwidth} X p{0.23\textwidth}}
\toprule
\textbf{ID} & \textbf{Paper name} & \textbf{Relation to questions} \\
\midrule
1  & On Designing Biopolymer-Bound Soil Composites (BSC) for Peak Compressive Strength \citep{bsc_1} & Directly related \\
2  & Development of biopolymer composites using lignin: A sustainable technology for fostering a green transition in the construction sector \citep{bsc_2} & Directly related \\
3  & Sustainability Assessment of Protein-Soil Composite Materials for Limited Resource Environments \citep{bsc_3} & Directly related \\
4  & Multiscale Modeling and Testing of Protein-Bound Regolith and Soils \citep{bsc_4} & Directly related \\
5  & Green recyclable biocomposite prepared from lignin and bamboo \citep{bsc_5} & General Material \\
6  & Life cycle assessment and design of LignoBlock: A lignin bound block on the path towards a green transition of the construction industry \citep{bsc_6} & Directly related \\
7  & Recycling of lignin-based biocomposites: Improving sustainability and enhancing material strength \citep{bsc_7} & Directly related \\
8  & Taking advantage of biological binders to solidify granular material: manufacture and recyclability of lignin-based biopolymer composites \citep{bsc_8} & Directly related \\
9  & Solvent selection enables sustainable and affordable lignin biocomposite for cement-free construction \citep{bsc_9} & Directly related \\
10 & Toward Cement-Free Construction: Microstructural Evaluation of Engineered Biopolymer Composite & Directly related \\
11 & Hypervelocity Impact Performance of Biopolymer-Bound Soil Composites for Space Construction \citep{bsc_11} & Not related to questions \\
12 & Determining the yield stress of a Biopolymer-bound Soil Composite for extrusion-based 3D printing applications \citep{bsc_12} & Not related to questions \\
13 & A shape stability model for 3D printable biopolymer-bound soil composite \citep{bsc_13} & Not related to questions \\
14 & Determining the structuration of biopolymer-bound soil composite \citep{bsc_14} & Not related to questions \\
15 & A Study on the Flow Behavior and Thixotropy of Biopolymer-Bound Soil Composite \citep{bsc_15} & Not related to questions \\
16 & BioSys: Efficient Quality Control System for Manufacturing of Sustainable Biopolymer Composites \citep{bsc_16} & Not related to questions \\
17 & AI-powered non-destructive testing for smart manufacturing of carbon-negative biopolymer-bound soil composite \citep{bsc_17} & Not related to questions \\
18 & Prediction of ultimate compressive strength for biopolymer-bound soil composites (BSC) using sliding wingtip crack analysis \citep{bsc_18} & Not related to questions \\
19 & Creation of Statistically Equivalent Periodic Unit Cells for Protein-Bound Soils \citep{bsc_19} & Not related to questions \\
20 & A comprehensive review of various biopolymer composites and their applications: From biocompatibility to self-healing \citep{bsc_20} & General Material \\
21 & Synthetic biology for fibers, adhesives, and active camouflage materials in protection and aerospace \citep{bsc_21} & General Material \\
22 & Blood, sweat, and tears: extraterrestrial regolith biocomposites with in vivo binders \citep{bsc_22} & Similar materials to BSC \\
23 & Effects of Xanthan gum biopolymer on soil strengthening \citep{bsc_23} & Similar materials to BSC \\
24 & Evaluating the physical characteristics of biopolymer/soil mixtures \citep{bsc_24} & Similar materials to BSC \\
25 & Geotechnical behavior of a beta-1,3/1,6-glucan biopolymer-treated residual soil \citep{bsc_25} & Similar materials to BSC \\
26 & Application of Microbial Biopolymers as an Alternative Construction Binder for Earth Buildings in Underdeveloped Countries \citep{bsc_26} & Similar materials to BSC \\
27 & Improving Mechanical Properties of Sand Using Biopolymers \citep{bsc_27} & Similar materials to BSC \\
28 & Mechanical properties of biopolymer-stabilised soil-based construction materials \citep{bsc_28} & Similar materials to BSC \\
29 & Seaweed biopolymers as additives for unfired clay bricks \citep{bsc_29} & Similar materials to BSC \\
30 & StarCrete: A starch-based biocomposite for off-world construction \citep{bsc_30} & Similar materials to BSC \\
\bottomrule
\end{tabularx}

\vspace{0.5\baselineskip}
\captionof{table}{Papers included in the corpus and their relation to the question sets.}
\label{tab:bsc_papers}
\end{center}

\clearpage
\twocolumn

\end{document}